\documentclass[runningheads]{llncs}
\usepackage{graphicx}

\usepackage{tikz}
\usepackage{comment}
\usepackage{amsmath,amssymb} 
\usepackage{color}

\usepackage{graphicx}
\usepackage{amsmath}
\usepackage{amssymb}
\usepackage{booktabs}
\usepackage{multirow}
\newcommand{\cmark}{\ding{51}}%
\usepackage{xcolor}
\newcommand{\fancomment}[1]{\textcolor[rgb]{1,0,0} {#1}}

\usepackage{comment}
\usepackage{amssymb}
\usepackage{pifont}
\usepackage{bbm}
\newcommand{\etal}{\textit{et al}.}
\newcommand{\ie}{\textit{i}.\textit{e}.}
\newcommand{\eg}{\textit{e}.\textit{g}.}
\newcommand\blfootnote[1]{%
  \begingroup
  \renewcommand\thefootnote{}\footnote{#1}%
  \addtocounter{footnote}{-1}%
  \endgroup
}

\usepackage[pagebackref,breaklinks,colorlinks]{hyperref}

\usepackage[capitalize]{cleveref}
\crefname{section}{Sec.}{Secs.}
\Crefname{section}{Section}{Sections}
\Crefname{table}{Table}{Tables}
\crefname{table}{Tab.}{Tabs.}

\usepackage[accsupp]{axessibility}  

\begin{document}
\pagestyle{headings}
\mainmatter
\def\ECCVSubNumber{1866}  

\title{Self-Support Few-Shot Semantic Segmentation} 

\titlerunning{Self-Support Few-Shot Semantic Segmentation}
%
\author{Qi Fan\inst{1}\and 
Wenjie Pei\inst{2}$^\dagger$\and 
Yu-Wing Tai\inst{1,3} \and
Chi-Keung Tang\inst{1}}
\authorrunning{Qi Fan et al.}
%
\institute{$^1$ HKUST, 
$^2$ Harbin Institute of Technology, Shenzhen,
$^3$ Kuaishou Technology\\
\email{fanqics@gmail.com, wenjiecoder@outlook.com, yuwing@gmail.com, cktang@cs.ust.hk}}
\maketitle

\begin{abstract}
   Existing few-shot segmentation methods have achieved great progress based on the support-query matching framework.
   But they still heavily suffer from the limited coverage of intra-class variations 
   from the few-shot supports provided.
   Motivated by 
   the simple Gestalt principle that pixels belonging to the same object are more similar than those to different objects of same class, we propose a novel self-support matching strategy to alleviate this problem, which uses query prototypes to match query features, where the query prototypes are collected from high-confidence query predictions.
   This strategy can effectively capture the consistent underlying characteristics of the query objects, and thus fittingly match query features.
   We also propose an adaptive self-support background prototype generation module and self-support loss to further facilitate the self-support matching procedure.
   Our self-support network 
   substantially improves the prototype quality, 
   benefits more improvement from stronger backbones and more supports,
   and achieves SOTA on multiple datasets. 
   Codes are at \url{https://github.com/fanq15/SSP}.\blfootnote{This research was supported by Kuaishou Technology, the Research Grant Council of the HKSAR (16201420), and NSFC fund (U2013210, 62006060).} \blfootnote{$^\dagger$Corresponding author.}
\end{abstract}

\section{Introduction}

Semantic segmentation has achieved remarkable advances tapping into deep learning networks~\cite{Goodfellow-et-al-2016,krizhevsky2012imagenet,he2016deep} and large-scale datasets such as~\cite{deng2009imagenet,benenson2019large,zhou2017scene}. However, current high-performing semantic segmentation methods rely heavily on laborious pixel-level annotations, %
which has expedited the recent development of 
few-shot semantic segmentation (FSS).


Few-shot semantic segmentation aims to segment arbitrary novel classes using only a few support samples.
The dilemma is that the support images are limited and fixed (usually $\{1,3,5,10\}$ supports per class), while the query images can be massive and arbitrary. Limited few-shot supports can easily fail to cover underlying appearance variations of the target class in query images, regardless of the support quality.
This is clearly caused by the inherent data scarcity and diversity, two long standing issues in few-shot learning.

Existing methods try to solve the problem by making full use of the limited supports, such as proposing better matching mechanism~\cite{siam2020weakly,liu2020dynamic,yang2020brinet,wang2020few,zhang2019pyramid,he2021progressive,zhuge2021deep} or generating representative prototypes~\cite{siam2019amp,li2021adaptive,ouyang2020self,liu2020part,yang2021mining,kim2021uncertainty,nguyen2019feature,yang2020prototype,zhang2019canet,gairola2020simpropnet}.
Despite their success, they still cannot fundamentally solve the appearance discrepancy problem, bounded by the scarce few-shot supports.

\begin{figure}[!t]
\centering
\includegraphics[width=1.0\linewidth]{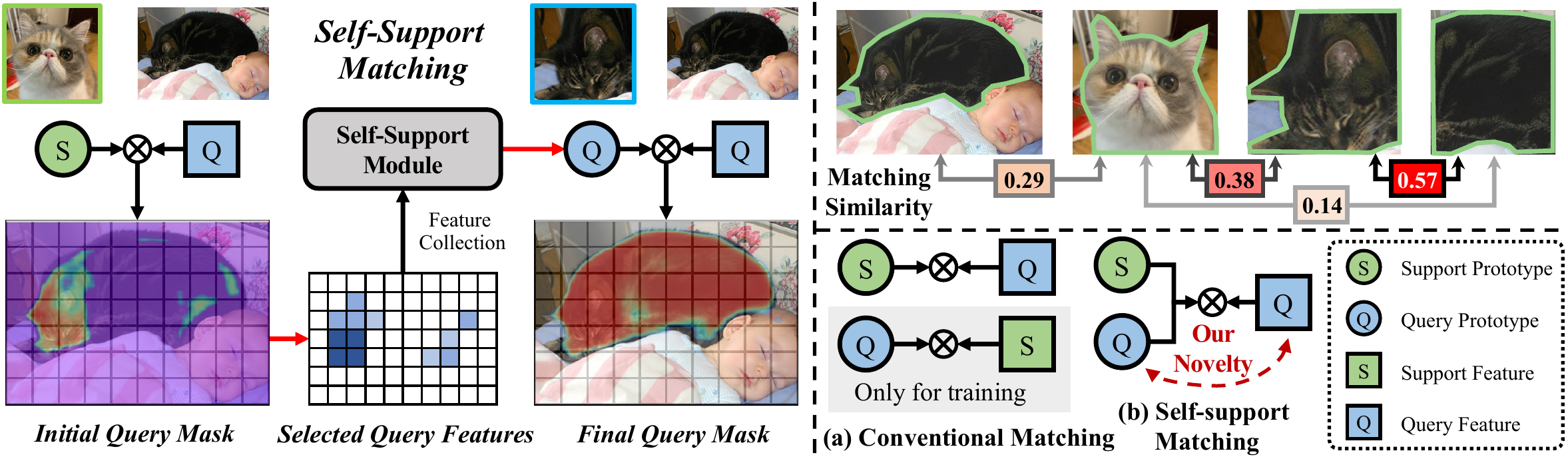}
\vspace{-0.2in}
\caption{The \textit{left} image illustrates the core idea of our self-support matching. We use the initial query mask prediction to collect query features in high-confidence regions and then use the generated \textbf{query prototype} to perform \textbf{self-matching} with \textbf{query features}.
The \textit{right top} image illustrates the motivation of our self-support matching: pixels/regions of the same objects are more similar than those from different objects. The numbers in boxes represent the cosine similarities between two objects.
The \textit{right bottom} image illustrates that our self-support matching is fundamentally distinct from conventional matching methods.}
\label{fig:teaser}
\end{figure}

We propose a novel self-support matching strategy to narrow  the matching appearance discrepancy.
This strategy uses query prototypes to match query features, or in other words, use the query feature to self-support itself. We thus call the query prototype as {\em self-support prototype} because of its self-matching property.
This new idea is motivated by the classical Gestalt law~\cite{gestalt} that pixels belonging to the same object are more similar than those to different objects.



Refer to Figure~\ref{fig:teaser} for a high-level understanding of our novel self-support matching.  First we generate the initial mask predictions by directly matching the support prototype and query features.
Based on the initial query mask, we collect confident query features to generate the self-support prototype, which is  used to perform matching with query features.
Our {\em self-support module (SSM)} collects confident features of the cat head which are used to segment the entire black cat.
Our model is optimized on base classes to retrieve other object parts supported by object fragments, \ie, self-support prototype.



We apply our self-support module on both foreground and background prototypes for self-support matching.
While SSM directly benefits foreground prototypes, 
note that the background is usually cluttered, which does not have the global semantic commonality shared among all background pixels.
Thus, rather than generating a global background prototype by aggregating all the background pixels, we propose to adaptively generate self-support background prototypes for each query pixel, by dynamically aggregating similar background pixels in the query image.
The {\em adaptive self-support background prototype (ASBP)} is motivated by the fact that separate background regions have local semantic similarity.
Finally, we propose a {\em self-support loss (SSL)} to further facilitate the self-support procedure.

Our self-support matching strategy is thus fundamentally different than conventional support-query matching. We use the flexible self-support prototypes to match query features, which can effectively capture the consistent underlying characteristics of the query objects, and thus fittingly match query features.
As shown in Figure~\ref{fig:teaser}, the cats in the query and support images are very different in color, parts and scales,
The garfield cat support has large appearance discrepancy to the black cat query, and undoubtedly conventional support-query matching produces inferior segmentation.  In our self-support matching, our self-support prototype (the black cat head) is more consistent to the query (the entire black cat), and thus our method produces satisfactory results.

We are the first to perform self-support matching between query prototype and query features.
As shown in Figure~\ref{fig:teaser}, 
our self-support matching fundamentally differs from conventional matching. Other methods learn better support prototypes for support-query matching from extra unlabeled images (PPNet~\cite{liu2020ppnet} and MLC~\cite{yang2021mining}) or builds various support prototype generation modules~\cite{siam2019amp,li2021adaptive,yang2020prototype} or feature priors (PFENet~\cite{tian2020prior}) based on support images.
Although PANet~\cite{wang2019panet} and CRNet~\cite{liu2020crnet} also explore query prototypes, they use \textit{query prototypes} to match \textit{support features} as a query-support matching only for  auxiliary training, and cannot solve the appearance discrepancy.

Our self-support method significantly improves the prototype quality by alleviating 
the intra-class appearance discrepancy problem,
evidenced by the performance boost on multiple datasets in our experimental validation. 
Despite the simple idea, our self-support method is very effective and has various advantages, such as benefiting more from stronger backbone and more supports, producing high-confidence predictions, more robustness to weak support labels, higher generalization to other methods and higher running efficiency. We will substantiate these advantages with thorough experiments. In summary, our contributions are:
\begin{itemize}
\setlength{\itemsep}{0.5pt}
\setlength{\parsep}{0.5pt}
\setlength{\parskip}{0.5pt}
    \item We propose  novel self-support matching and build a novel self-support network to solve the appearance discrepancy problem in FSS.
    \item We propose self-support prototype, adaptive self-support background prototype and self-support loss to facilitate our self-support method.
    \item Our self-support method benefits more improvement from stronger backbones and more supports, and outperforms previous SOTAs on multiple datasets with many desirable advantages.
\end{itemize}

\section{Related Works}

\noindent{\bf Semantic Segmentation.}
Semantic segmentation is a fundamental computer vision task to produce pixel-wise dense semantic predictions. The state-of-the-art has recently been greatly advanced by the end-to-end fully convolutional network (FCN)~\cite{long2015fully}. Subsequent works have since followed this FCN paradigm and contributed many effective modules to further promote the performance, such as encoder-decoder architectures~\cite{badrinarayanan2017segnet,cheng2019spgnet,chen2018encoder,ronneberger2015u}, image and feature pyramid modules~\cite{kirillov2019panoptic,zhao2017pyramid,chen2016attention,lin2017refinenet,lin2016efficient}, context aggregation modules~\cite{fu2019dual,fu2019adaptive,he2019adaptive,zhao2018psanet,zhu2019asymmetric,huang2019ccnet,yuan2020object,zhang2019acfnet} and advance convolution layers~\cite{yu2017dilated,chen2017deeplab,dai2017deformable,noh2015learning}.
Nevertheless, the above  segmentation methods rely heavily on abundant pixel-level annotations. This paper aims to  tackle the semantic segmentation problem in the few-shot scenario.

\noindent{\bf Few-Shot Learning.}
Few-shot learning targets at recognizing new concepts from very few samples. This low cost property has attracted a lot of  research interests over the last years. There are three main approaches.
The first is the transfer-learning approach~\cite{chen2019closer,gidaris2018dynamic,dhillon2019baseline,qi2018low} by adapting the prior knowledge learned from base classes to novel classes in a two-stage finetuning procedure. 
The second is the optimized-based approach~\cite{Finn2017ModelAgnosticMF,bertinetto2018meta,lee2018gradient,gordon2018metalearning,lee2019meta,antoniou2018train,grant2018recasting,rusu2018meta}, which rapidly updates models through meta-learning the optimization procedures from a few samples.
The last is the metric-based approach~\cite{allen2019infinite,doersch2020crosstransformers,hou2019cross,koch2015siamese,li2019finding,li2019revisiting}, which applies a siamese network~\cite{koch2015siamese} on support-query pairs to learn a general metric for evaluating their relevance. 
Our work, including many few-shot works~\cite{fan2020few,kang2018few,zhang2020few,yan2019metarcnn,fan2021few} on various high-level computer vision tasks, are inspired by the metric-based approach.

\noindent{\bf Few-Shot Semantic Segmentation.}
Few-shot semantic segmentation is pioneered by Shaban \etal~\cite{shaban2017one}.
Later works have mainly adopted the metric-based mainstream paradigm~\cite{dong2018few} 
with various improvements, \eg, improving the matching procedure between support-query images with various attention mechanisms~\cite{siam2020weakly,liu2020dynamic,yang2020brinet}, better optimizations~\cite{zhu2020self,liu2021learning}, memory modules~\cite{wu2021learning,xie2021few}, graph neural networks~\cite{xie2021scale,wang2020few,zhang2019pyramid}, learning-based classifiers~\cite{tian2020differentiable,lu2021simpler}, progressive matching~\cite{he2021progressive,zhuge2021deep}, or other advanced techniques~\cite{zhang2021self,liu2021anti,min2021hypercorrelation,li2020fss}.

We are the first to perform self-support matching between query prototype and query features.
Our self-support matching method is also related to the prototype generation methods. 
Some methods leverage extra unlabeled data~\cite{yang2021mining,liu2020ppnet} or feature priors~\cite{tian2020prior} for further feature enhancement.
Other methods 
generate representative support prototypes with various techniques, \eg, 
attention mechanism~\cite{zhang2019canet,gairola2020simpropnet}, adaptive prototype learning~\cite{siam2019amp,li2021adaptive,ouyang2020self}, or various prototype generation approaches~\cite{nguyen2019feature,yang2020prototype}.
Although the query prototype has been explored in some methods~\cite{wang2019panet,liu2020crnet}, they only use query prototypes to match support features for prototype regularization. 
Finally, existing methods heavily suffer from the intra-class discrepancy problem in the support-query matching. On the other hand, 
we propose a novel self-support matching strategy to effectively address this matching problem.


\section{Self-Support Few-Shot Semantic Segmentation}

Given only a few support images, few-shot semantic segmentation aims to segment objects of novel classes using the model generalized from base classes. 
Existing mainstream few-shot semantic segmentation solution can be formulated as follows: 
The input support and query images $\{I_s, I_q\}$ are processed by a weight-shared backbone to extract image features $\{\mathcal{F}_s, \mathcal{F}_q\} \in \mathbbm{R}^{C \times H \times W}$, where $C$ is the channel size and $H \times W$ is the feature spatial size. Then the support feature $\mathcal{F}_s$ and its groundtruth mask $\mathcal{M}_s$ are fed into the masked average pooling layer to generate the support prototype vectors $\mathcal{P}_s = \{\mathcal{P}_{s,f}, \mathcal{P}_{s,b}\} \in \mathbbm{R}^{C \times 1 \times 1}$ for foreground and background regions respectively. Finally, two distance maps $\mathcal{D}=\{\mathcal{D}_{f}, \mathcal{D}_{b}\}$ are generated by evaluating the cosine similarity between $\mathcal{P}_s$ and $\mathcal{F}_q$, 
which is then processed by a softmax operation as the final prediction ${\mathcal M}_1=\text{softmax}({\mathcal D})$.

\subsection{Motivation}

Current FSS methods rely heavily on the support prototype to segment query objects, by densely matching each query pixel with the support prototype.
However, such cross-object matching severely suffers  from  intra-class appearance discrepancy, where objects in support and query can look very different 
even belonging to the same class.
Such high intra-class variation cannot be 
reconciled by only a few supports, thus leading to poor matching results due to the large appearance gap between the query and supports.

To validate the relevance of Gestalt law~\cite{gestalt} in narrowing such appearance discrepancy, we statistically analyze the feature cosine similarity of cross-object and intra-object pixels of Pascal VOC~\cite{everingham2010pascal}, where the pixel features are extracted from the ImageNet~\cite{deng2009imagenet}-pretrianed ResNet-50~\cite{he2016deep}.
Table~\ref{table:teaser-1} shows that pixels belonging to the same object are much more similar than the cross-object pixels. Notably, background pixels share similar characteristics on their own, where intra-image background pixels are much more similar than cross-image pixels.

Thus, we propose to leverage the query feature to generate self-support prototypes to match the query feature itself.
Notably, such prototype aligns 
the query along the homologous query features and thus can significantly narrow  the feature gap between the support and query. In hindsight, 
the crucial reason 
the self-support matching works better than traditional support-query matching is that for a given visual object class, the intra-object similarities are much higher than the cross-object similarities. 

\subsection{Self-Support Prototype}

\begin{table}[!t]
\begin{center}
\tabcolsep=8pt 
\caption{Cosine similarity for cross/intra object pixels.}
\vspace{-0.1in}
\begin{tabular}{cc|cc}
\toprule
\multicolumn{2}{c|}{FG Pixels Similarity} & \multicolumn{2}{c}{BG Pixels Similarity} \\
\midrule
cross-object & intra-object & cross-image & intra-image  \\
0.308 & 0.416$_{\uparrow 0.108}$ & 0.298 & 0.365$_{\uparrow 0.067}$ \\
\bottomrule
\end{tabular}
\label{table:teaser-1}
\vspace{-0.3in}
\end{center}
\end{table}

\begin{figure*}[!t]
\centering
\includegraphics[width=1.0\linewidth]{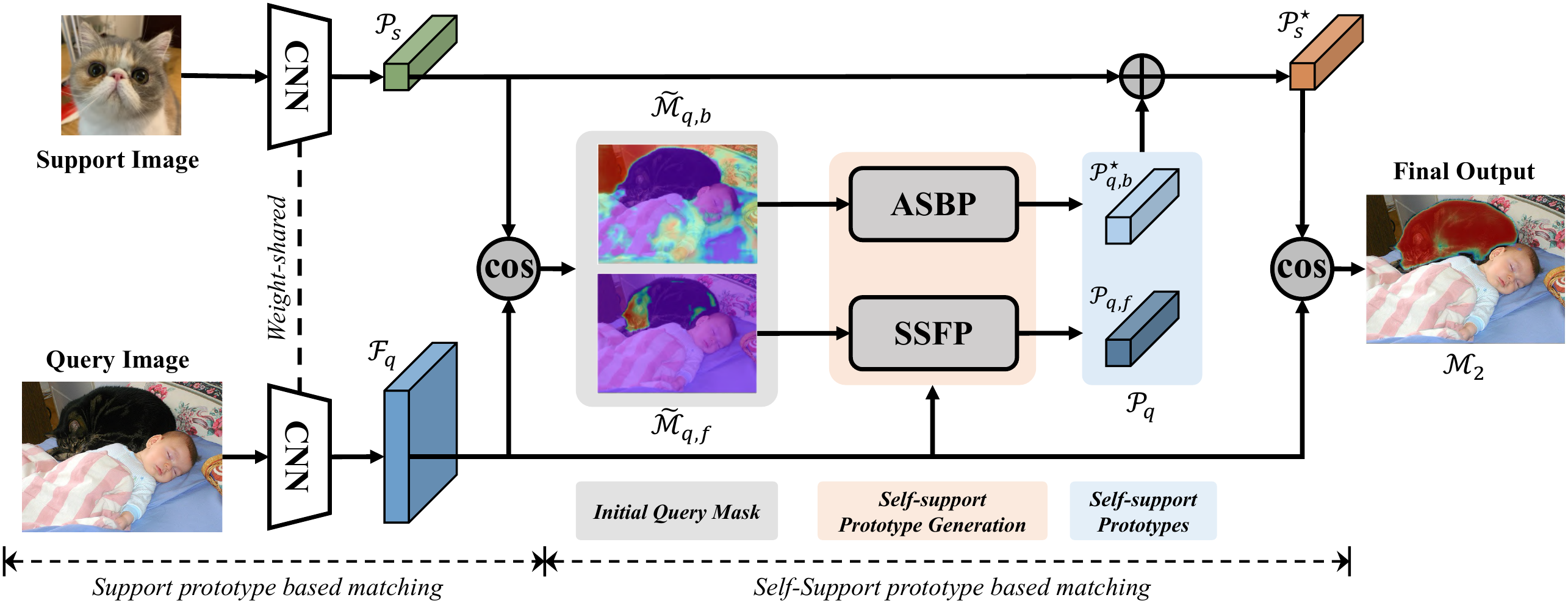}
\vspace{-0.2in}
\caption{Overall self-support network architecture. We first generate the initial mask predictions using the traditional support prototype based matching network.
Then we leverage the initial query mask to aggregate query features to generate self-support prototypes, \ie, the self-support foreground prototype (SSFP) and adaptive self-support background prototype (ASBP). Finally, we combine the support prototype and self-support prototypes to perform matching with query features.} 
\label{fig:network}
\vspace{-0.1in}
\end{figure*}

Our core idea (Figure~\ref{fig:network}) is to aggregate query features to generate the query prototype and use it to self-support the query feature itself.

To recap, the regular support prototype generation procedure is: 
\begin{equation}
\setlength{\abovedisplayskip}{3pt}
\setlength{\belowdisplayskip}{3pt}
    \mathcal{P}_s = 
    \mathit{MAP}(\mathcal{M}_s, \mathcal{F}_s),
\end{equation}
where $\mathit{MAP}$ is the masked average pooling operation, which is 
used to generate the matching prediction with query feature $\mathcal{F}_q$:
\begin{equation}
\setlength{\abovedisplayskip}{3pt}
\setlength{\belowdisplayskip}{3pt}
\label{eq:m1}
\mathcal{M}_1 = \text{softmax}(\text{cosine}(\mathcal{P}_s, \mathcal{F}_q)),
\end{equation}
where $\text{cosine}$ is the cosine similarity metric.

Now, we can generate the query prototype $\mathcal{P}_q$ in the same manner, except the groundtruth masks of query images $\mathcal{M}_q$ are unavailable during inference. Thus, we need to use a predicted query mask $\widetilde{\mathcal{M}}_q$ to aggregate query features. 
The query prototype generation procedure can be formulated as:
\begin{equation}
\setlength{\abovedisplayskip}{3pt}
\setlength{\belowdisplayskip}{3pt}
\label{eq:pq}
    \mathcal{P}_q = \mathit{MAP}(\widetilde{\mathcal{M}}_q, \mathcal{F}_q),
\end{equation}
where $\widetilde{\mathcal{M}}_q=\mathbbm{1}(\mathcal{M}_1 > \tau)$, and $\mathcal{M}_1$ is the estimated query mask generated by Equation~\ref{eq:m1}, $\mathbbm{1}$ is the indicator function. The mask threshold $\tau$ is used to control the query feature sampling scope which is set as $\{\tau_{fg}=0.7, \tau_{bg}=0.6\}$ for foreground and background query masks respectively. 
The estimated self-support prototype $\mathcal{P}_q=\{\mathcal{P}_{q,f}, \mathcal{P}_{q,b}\}$ will be utilized to match query features.

We understand the reader's natural concern about the quality of self-support prototype, which is generated based on the estimated query mask, \ie, whether the estimated mask is capable of effective self-support prototype generation.
We found that even the estimated query mask is not perfect, as long as it covers some representative object fragments, it is sufficient to retrieve other regions of the same object.
To validate  partial object or object fragment is capable of supporting the entire object, we train and evaluate models with partial prototypes, which are aggregated from randomly selecting features based on the groundtruth mask labels.
We conduct the 1-shot segmentation experiments on Pascal VOC dataset with the ResNet-50 backbone.
As shown in Table~\ref{table:teaser-2}, while reducing the aggregated object regions for prototype generation, our self-support prototype consistently achieves high segmentation performance. By contrast, the traditional support prototype consistently obtains much inferior performance, even using  perfect support features from the entire object.

\begin{table}[!t]
\begin{center}
\tabcolsep=8pt 

\caption{The 1-shot matching results (mIoU) of support/self-support prototypes aggregated from full/partial objects.}
\vspace{-0.1in}

\begin{tabular}{ccccc}
\toprule
Object Ratio            & full & $10\%$ & $1\%$ & $1\%$+noise \\ 
\midrule
Support Prototype      & 58.2    &   57.1     & 52.4    & 48.7 \\ 
Self-support Prototype & 83.0    &   82.5     & 79.2    & 74.6 \\ 
\bottomrule
\end{tabular}

\label{table:teaser-2}
\end{center}
\vspace{-0.1in}

\end{table}

We further introduce noisy features (with $20\%$ noise ratio) 
into partial prototypes to mimic realistic self-support generation during inference, by randomly selecting image features from non-target regions and aggregating these features into the above partial prototypes.
To our pleasant surprise, our self-support prototype still works much better than the traditional support prototype in such noisy situation.
Note that each image may contain multiple objects, thus the good performance indicates that our self-support prototype can also handle well the multiple objects scenarios.
These results confirm the practicability and advantages of our self-support prototypes in the realistic applications.

\begin{figure*}[!t]
\centering
\includegraphics[width=1.0\linewidth]{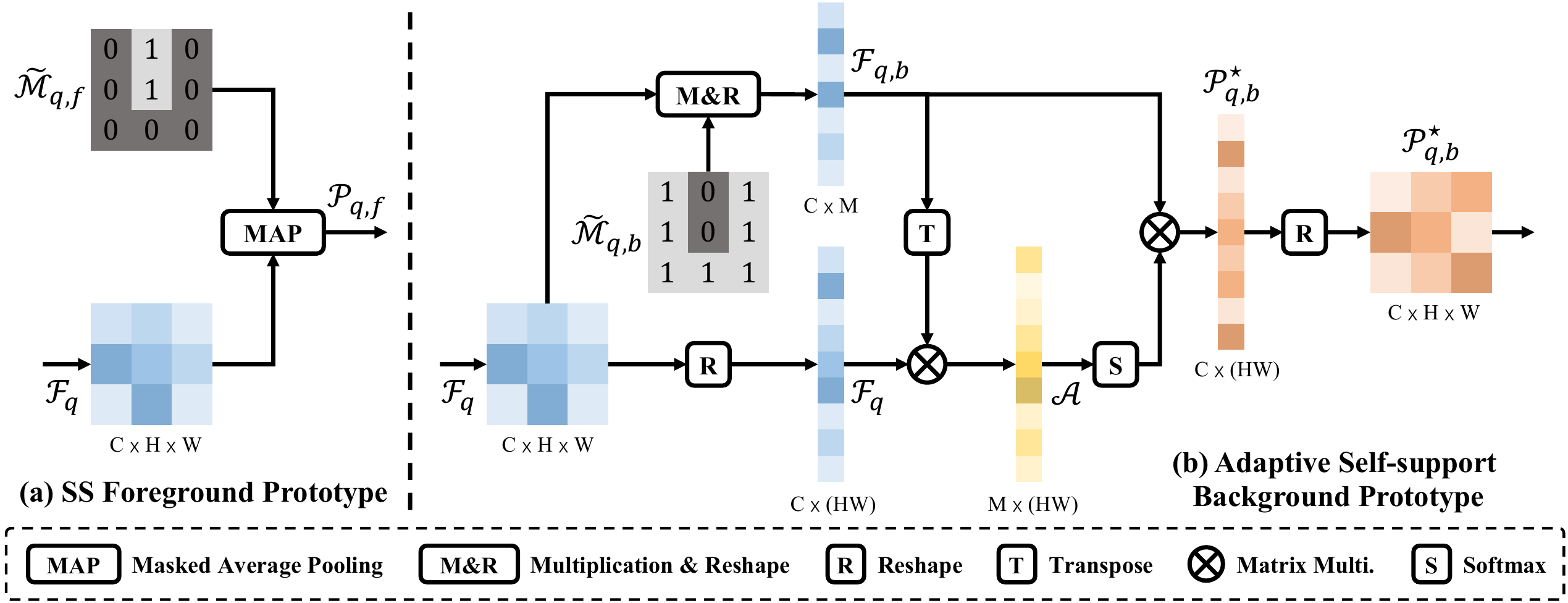}
\vspace{-0.2in}
\caption{Prototype generations of (a) self-support (SS) foreground prototype and (b) adaptive self-support background prototype.} 
\vspace{-0.1in}
\label{fig:pg}
\end{figure*}

\subsection{Adaptive Self-Support Background Prototype}



Foreground pixels share semantic commonalities~\cite{fan2021group,fan2020cpmask}, which constitutes the rationale behind our self-support prototype generation and matching procedure between query feature and support prototypes for foreground objects.
Therefore, we can utilize a masked average pooling to generate the self-support foreground prototype (Figure~\ref{fig:pg} (a)):
\begin{equation}
\setlength{\abovedisplayskip}{3pt}
\setlength{\belowdisplayskip}{3pt}
    \mathcal{P}_{q,f} = \mathit{MAP}(\widetilde{\mathcal{M}}_{q,f}, \mathcal{F}_q),
\end{equation}
where $\widetilde{\mathcal{M}}_{q,f}$ is the aforementioned estimated query mask.

On the other hand, background can be  cluttered, where commonalities can be reduced to local semantic similarities in disjoint regions, without a global semantic commonality shared among all background pixels. For example, for a query image with dog as the target class, other objects such as person and car are both treated as background, but they are different in both appearance and semantic levels.
This observation is also validated by the smaller background pixel similarity compared to foreground pixels as shown in Table~\ref{table:teaser-1}, especially in the intra-object/image situation. 
This motivates us to generate multiple self-support background prototypes for different query semantic regions.

A straightforward solution is to directly group multiple background prototypes using a clustering algorithm, and then choose the most similar prototype at each query pixel for background matching. 
This explicit background grouping heavily relies on the clustering algorithm, which is unstable and time-consuming.
Therefore, we propose a more flexible and efficient method to adaptively generate self-support background prototypes for each query pixel (Figure~\ref{fig:pg} (b)).

The idea is to dynamically aggregate similar background pixels for each query pixel to generate adaptive self-support background prototypes.
Specifically, we first gather the background query features $\mathcal{F}_{q,b} \in \mathbbm{R}^{C \times M}$ through the masked multiplication on the query feature $\mathcal{F}_q$ with the background mask $\widetilde{\mathcal{M}}_{q,b}$, where $M$ is the pixel number of the background region. Then we can generate the affinity matrix $\mathcal{A}$  between pixels of the reshaped background query feature $\mathcal{F}_{q,b}$ and full query feature $\mathcal{F}_q$ through a matrix multiplication operation $\mathit{MatMul}$:
\begin{equation}
\setlength{\abovedisplayskip}{3pt}
\setlength{\belowdisplayskip}{3pt}
    \mathcal{A} = \mathit{MatMul}({\mathcal{F}_{q,b}}^T, \mathcal{F}_q),
\end{equation}
where $\mathcal{A}$ is in size of $\mathbbm{R}^{M \times (H \times W)}$. 
The affinity matrix is normalized through a softmax operation along the first dimension, which is used to weighted aggregate background query features for each query pixel to generate the adaptive self-support background prototypes $\mathcal{P}_{q,b}^\star \in \mathbbm{R}^{C \times H \times W}$:
\begin{equation}
\setlength{\abovedisplayskip}{3pt}
\setlength{\belowdisplayskip}{3pt}
    \mathcal{P}_{q,b}^\star = \mathit{MatMul}(\mathcal{F}_{q,b}, \text{softmax}(\mathcal{A})).
\end{equation}
The self-support prototype is updated with the adaptive self-support background prototype: $\mathcal{P}_{q} = \{\mathcal{P}_{q,f}, \mathcal{P}_{q,b}^\star\}$.

\subsection{Self-Support Matching}

We weighted combine the support prototype 
$\mathcal{P}_s$
and self-support prototype $\mathcal{P}_q$:
\begin{equation}
\setlength{\abovedisplayskip}{3pt}
\setlength{\belowdisplayskip}{3pt}
    \mathcal{P}_s^\star = \alpha_1 \mathcal{P}_s + \alpha_2 \mathcal{P}_q,
\end{equation}
where $\alpha_1$ and $\alpha_2$ are the tuning weights and we set $\alpha_1 = \alpha_2 = 0.5$ in our experiments.
Then we compute the cosine distance between the augmented support prototype $\mathcal{P}_s^\star$ and query feature $\mathcal{F}_q$ to generate the final matching prediction:
\begin{equation}
\setlength{\abovedisplayskip}{3pt}
\setlength{\belowdisplayskip}{3pt}
\label{eq:m2}
\mathcal{M}_2 = \text{softmax}(\text{cosine}(\mathcal{P}_s^\star, \mathcal{F}_q)).
\end{equation}

Then we apply the training supervision on the generated distance maps: 
\begin{equation}
\label{eq:cross_loss}
\setlength{\abovedisplayskip}{3pt}
\setlength{\belowdisplayskip}{3pt}
    \mathcal{L}_{m} = \mathit{BCE}(\text{cosine}(\mathcal{P}_s^\star, \mathcal{F}_q), \mathcal{G}_q),
\end{equation}
where $\mathit{BCE}$ is the binary cross entropy loss and $\mathcal{G}_q$ is the groundtruth mask of the query image.

To further facilitate  the self-support matching procedure, we propose a novel query self-support loss. For the query feature $\mathcal{F}_q$ and its prototype $\mathcal{P}_q$, we apply the following training supervision:
\begin{equation}
\setlength{\abovedisplayskip}{3pt}
\setlength{\belowdisplayskip}{3pt}
    \mathcal{L}_{q} = \mathit{BCE}(\text{cosine}(\mathcal{P}_q, \mathcal{F}_q), \mathcal{G}_q).
\end{equation}
We can apply the same procedure on the support feature to introduce the support self-matching loss $\mathcal{L}_{s}$. 

Finally, we train the model in an end-to-end manner by jointly optimizing all the aforementioned losses:
\begin{equation}
\setlength{\abovedisplayskip}{3pt}
\setlength{\belowdisplayskip}{3pt}
    \mathcal{L} = \lambda_1 \mathcal{L}_{m} + \lambda_2 \mathcal{L}_{q} + \lambda_3 \mathcal{L}_{s},
\end{equation}
where $\lambda_1=1.0, \lambda_2=1.0, \lambda_3=0.2$ are the loss weights.

\section{Experiments}

\noindent{\bf Datasets.~} We conduct experiments on two FSS benchmark datasets: PASCAL-5$^i$~\cite{everingham2010pascal} and COCO-20$^i$~\cite{lin2014microsoft}. 
We follow previous works~\cite{tian2020prior,yang2021mining} to split the data into four folds for cross validation, where three folds are used for training and the remaining one for evaluation. During inference, we randomly sample 1,000/4,000 support-query pairs to perform evaluation for PASCAL-5$^i$ and COCO-20$^i$, respectively. We use the popular mean Intersection-over-Union (mIoU, $\uparrow$\footnote{The ``$\uparrow$'' (``$\downarrow$'') means that the higher (lower) is better.}) as the default metric to evaluate our model under 1-shot and 5-shot settings. We also apply the Mean Absolute Error (MAE, $\downarrow$) to evaluate our prediction quality. By default, all analyses are conducted on PASCAL-5$^i$ dataset with ResNet-50 backbone in the 5-shot setting.

\noindent{\bf Implementation details.~} We adopt the popular ResNet-50/101~\cite{he2016deep} pretrained on ImageNet~\cite{deng2009imagenet} as the backbone. Following previous work MLC~\cite{yang2021mining}, we discard the last backbone stage and the last ReLU for better generalization.
We use SGD to optimize our model with the 0.9 momentum and 1e-3 initial learning rate, which decays by 10 times every 2,000 iterations. The model is trained for 6,000 iterations where each training batch contains 4 support-query pairs. Both images and masks are resized and cropped into (473, 473) and augmented with random horizontal flipping. 
The evaluation is performed on the original image.

\begin{table*}[!t]
\begin{center}
\caption{Quantitative comparison results on PASCAL-$5^i$ dataset. The {\bf best} and \underline{second best} results are highlighted with {\bf bold} and \underline{underline}, respectively.}
\vspace{-0.1in}
\scalebox{0.93}{\begin{tabular}{l|c|ccccc|ccccc|c}
\toprule
& & \multicolumn{5}{|c|}{1-shot} & \multicolumn{5}{|c|}{5-shot} & \\
Method  & Backbone & fold0 & fold1 & fold2 & fold3 & {\bf Mean} & fold0 & fold1 & fold2 & fold3 &{\bf Mean} & Params \\
\midrule
PANet~\cite{wang2019panet}   & \multirow{8}{*}{Res-50} & 44.0 & 57.5 & 50.8 & 44.0 & 49.1 & 55.3 & 67.2 & 61.3 & 53.2 & 59.3 & \underline{23.5 M} \\
PPNet~\cite{liu2020part}   & & 48.6 & 60.6 & 55.7 & 46.5 & 52.8 & 58.9 & 68.3 & 66.8 & 58.0 & 63.0 & 31.5 M \\
PFENet~\cite{tian2020prior}  & & \underline{61.7} & 69.5 & 55.4 & \underline{56.3} & 60.8 & 63.1 & 70.7 & 55.8 & 57.9 & 61.9 & 34.3 M \\
CWT~\cite{lu2021simpler}         & & 56.3 & 62.0 & 59.9 & 47.2 & 56.4 & 61.3 & 68.5 & 68.5 & 56.6 & 63.7 & - \\
HSNet~\cite{min2021hypercorrelation}         & & {\bf 64.3} & \underline{70.7} & 60.3 & {\bf 60.5} & {\bf 64.0} & {\bf 70.3} & {\bf 73.2} & 67.4 & {\bf 67.1} & {\bf 69.5} & 26.1 M \\
MLC~\cite{yang2021mining}         & & 59.2 & {\bf 71.2} & \underline{65.6} & 52.5 & \underline{62.1} & 63.5 & 71.6 & 71.2 & 58.1 & 66.1 & {\bf 8.7 M} \\
SSP (Ours)       & & 61.4 & 67.2 & 65.4 & 49.7 & 60.9 & \underline{68.0} & 72.0 & \underline{74.8} & 60.2 & 68.8 & {\bf 8.7 M} \\
SSP$_{\mathit{refine}}$       & & 60.5 & 67.8 & {\bf 66.4} & 51.0 & 61.4 & 67.5 & \underline{72.3} & {\bf 75.2} & \underline{62.1} & \underline{69.3} & {\bf 8.7 M} \\
\midrule
\midrule
FWB~\cite{nguyen2019feature}    & \multirow{8}{*}{Res-101} & 51.3 & 64.5 & 56.7 & 52.2 & 56.2 & 54.8 & 67.4 & 62.2 & 55.3 & 59.9 & \underline{43.0 M} \\
PPNet~\cite{liu2020part}   & & 52.7 & 62.8 & 57.4 & 47.7 & 55.2 & 60.3 & 70.0 & 69.4 & 60.7 & 65.1 & 50.5 M \\
PFENet~\cite{tian2020prior}  & & 60.5 & 69.4 & 54.4 & 55.9 & 60.1 & 62.8 & 70.4 & 54.9 & 57.6 & 61.4 & 53.4 M \\
CWT~\cite{lu2021simpler}         & & 56.9 & 65.2 & 61.2 & 48.8 & 58.0 & 62.6 & 70.2 & 68.8 & 57.2 & 64.7 & - \\
HSNet~\cite{min2021hypercorrelation}         & & {\bf 67.3} & {\bf 72.3} & 62.0 & {\bf 63.1} & {\bf 66.2} & {\bf 71.8} & 74.4 & 67.0 & {\bf 68.3} & 70.4 & 45.2 M \\
MLC~\cite{yang2021mining}         & & 60.8 & \underline{71.3} & 61.5 & \underline{56.9} & 62.6 & 65.8 & 74.9 & 71.4 & 63.1 & 68.8 & {\bf 27.7 M} \\
SSP (Ours) & &  \underline{63.7} & 70.1 & \underline{66.7} & 55.4 & 64.0 & 70.3 & \underline{76.3} & \underline{77.8} & 65.5 & \underline{72.5} & {\bf 27.7 M} \\
SSP$_{\mathit{refine}}$ & & 63.2 & 70.4 & {\bf 68.5} & 56.3 & \underline{64.6} & \underline{70.5} & {\bf 76.4} & {\bf 79.0} & \underline{66.4} & {\bf 73.1} & {\bf 27.7 M} \\
\bottomrule
\end{tabular}
}
\label{table:pascal}
\end{center}
\vspace{-0.3in}
\end{table*}

\subsection{Comparison with State-of-the-Arts}

To validate the effectiveness of our method, we conduct extensive comparisons with SOTA methods under different backbone networks and few-shot settings.

\noindent{\bf PASCAL-5$^i$.~} 
We present the results of our self-support method and the improved version with one extra self-support refinement.
As shown in Table~\ref{table:pascal},
our method substantially outperforms MLC~\cite{yang2021mining} by a large margin in the 5-shot setting, with the improvement jumping from 2.7\% to 3.7\% with the ResNet-50 backbone replaced by the stronger ResNet-101 network.
In the 1-shot setting, our slightly inferior performance is remedied by using the stronger ResNet-101 backbone, where we surpass MLC~\cite{yang2021mining} by 1.4\% improvement.
We can further promote the overall performance on PASCAL-5$^i$ up to 73.1\% with the self-support refinement, which is a simple and straightforward extension by repeating the self-support procedure.
It surpasses the previous SOTA~\cite{min2021hypercorrelation} by 2.7\%.
Note that our method is non-parametric and thus our model uses fewest parameters while achieving the best performance. 

\begin{table*}[!t]
\begin{center}
\caption{Quantitative comparison results on COCO-$20^i$ dataset. $^\star$ denotes the results are evaluated on the HSNet's evaluation protocol.}
\vspace{-0.1in}
\scalebox{0.93}{\begin{tabular}{l|c|ccccc|ccccc|c}
\toprule
& & \multicolumn{5}{|c|}{1-shot} & \multicolumn{5}{|c|}{5-shot} & \\
Method  & Backbone & fold0 & fold1 & fold2 & fold3 & {\bf Mean} & fold0 & fold1 & fold2 & fold3 & {\bf Mean} & Params \\
\midrule
PANet~\cite{wang2019panet}  & \multirow{7}{*}{Res-50} & 31.5 & 22.6 & 21.5 & 16.2 & 23.0 & 45.9 & 29.2 & 30.6 & 29.6 & 33.8 & \underline{23.5 M} \\
PPNet~\cite{liu2020part}  & & 36.5 & 26.5 & 26.0 & 19.7 & 27.2 & 48.9 & 31.4 & 36.0 & 30.6 & 36.7 & 31.5 M \\
CWT~\cite{lu2021simpler}        & & 32.2 & 36.0 & 31.6 & 31.6 & 32.9 & 40.1 & 43.8 & 39.0 & 42.4 & 41.3 & -  \\
MLC~\cite{yang2021mining}         & & {\bf 46.8} & 35.3 & 26.2 & 27.1 & 33.9 & {\bf 54.1} & 41.2 & 34.1 & 33.1 & 40.6 & {\bf 8.7 M} \\
SSP (Ours)       & & \underline{46.4} & 35.2 & 27.3 & 25.4 & 33.6 & \underline{53.8} & 41.5 & 36.0 & 33.7 & 41.3 & {\bf 8.7 M} \\
\cmidrule{1-1} \cmidrule{3-13}
HSNet$^\star$~\cite{min2021hypercorrelation}         &  & 36.3 & {\bf 43.1} & {\bf 38.7} & {\bf 38.7} & {\bf 39.2} & 43.3 & {\bf 51.3} & {\bf 48.2} & {\bf 45.0} & {\bf 46.9} & 26.1 M \\
SSP$^\star$ (Ours)       & & 35.5 & \underline{39.6} & \underline{37.9} & \underline{36.7} & \underline{37.4} & 40.6 & \underline{47.0} & \underline{45.1} & \underline{43.9} & \underline{44.1} & {\bf 8.7 M} \\
\midrule
\midrule
PMMs~\cite{yang2020prototype}    & \multirow{6}{*}{Res-101} & 29.5 & 36.8 & 28.9 & 27.0 & 30.6 & 33.8 & 42.0 & 33.0 & 33.3 & 35.5 & \underline{38.6 M} \\
CWT~\cite{lu2021simpler}          & & 30.3 & 36.6 & 30.5 & 32.2 & 32.4 & 38.5 & 46.7 & 39.4 & 43.2 & 42.0 & -  \\
MLC~\cite{yang2021mining}          & & \underline{50.2} & 37.8 & 27.1 & 30.4 & 36.4 & \underline{57.0} & 46.2 & 37.3 & 37.2 & 44.4 & {\bf 27.7 M} \\
SSP (Ours)        & & {\bf 50.4} & 39.9 & 30.6 & 30.0 & 37.7 & {\bf 57.8} & 47.0 & 40.2 & 39.9 & 46.2 & {\bf 27.7 M} \\
\cmidrule{1-1} \cmidrule{3-13}
HSNet$^\star$~\cite{min2021hypercorrelation}         &  & 37.2 & \underline{44.1} & \underline{42.4} & {\bf 41.3} & \underline{41.2} & 45.9 & \underline{53.0} & {\bf 51.8} & \underline{47.1} & \underline{49.5} & 45.2 M \\
SSP$^\star$ (Ours)        & & 39.1 & {\bf 45.1} & {\bf 42.7} & \underline{41.2} & {\bf 42.0} & 47.4 & {\bf 54.5} & \underline{50.4} & {\bf 49.6} & {\bf 50.2} & {\bf 27.7 M} \\
\bottomrule
\end{tabular}
}
\label{table:coco}
\end{center}
\vspace{-0.3in}

\end{table*}

\noindent{\bf COCO-20$^i$.~}
This is a very challenging dataset whose images usually contain multiple objects against a complex background. As shown in Table~\ref{table:coco}, our method obtains comparable or best results with the ResNet-50 backbone. When equipped with the stronger ResNet-101 backbone, our method significantly outperforms MLC~\cite{yang2021mining} with 1.3/1.8\% improvements in 1/5-shot settings.
To fairly compare to HSNet~\cite{min2021hypercorrelation}, we adopt their evaluation protocol to evaluate our method. Our method achieves SOTA when using the ResNet-101 backbone.
Our method also performs best on FSS-1000~\cite{li2020fss}, shown in the supplementary material.

\noindent{Note that our method benefits more improvement from stronger backbones and more supports because they provide better self-support prototypes, which will be validated later in Table~\ref{table:benefit}.}

\subsection{Ablation Studies}

As shown in Table~\ref{table:ablation}, our self-support module significantly improves the performance by 2.5 \%.
The self-support loss further facilitates the self-support procedure and promotes the performance to 68.1\%. The baseline model also benefits from the extra supervision of self-support loss.
After equipped with the adaptive self-support background prototype, the self-support module can obtain extra 0.9\% gain.
Integrating all modules, our self-support method significantly improves the performance from 64.8\% to 68.8\% based on the strong baseline.

\begin{table}[!t]
\begin{center}
\tabcolsep=8pt
\caption{Self-support model ablation results. ``SSM'' denotes the self-support module (containing the self-support foreground/background prototypes) , ``SSL'' denotes the self-support loss and ``ASBP'' denotes the adaptive self-support background prototype.}
\vspace{-0.1in}
\begin{tabular}{cccccccl}
\toprule
SSM & SSL & ASBP & fold0 & fold1 & fold2 & fold3 & {\bf Mean} \\
\midrule
       &        &        & 62.2 & 70.5 & 70.7 & 55.7 & 64.8 \\
\cmark &        &        & 65.3 & 71.1 & 73.6 & 59.2 & 67.3$_{\uparrow 2.5}$ \\
       & \cmark &        & 63.6 & 71.0 & 71.7 & 56.3 & 65.7$_{\uparrow 0.9}$ \\
\cmark & \cmark &        & 67.0 & {\bf 72.4} & 72.9 & 59.9 & 68.1$_{\uparrow 3.3}$ \\
\cmark &        & \cmark & 67.0 & 71.4 & 74.7 & 59.8 & 68.2$_{\uparrow 3.4}$ \\
\cmark & \cmark & \cmark & {\bf 68.0} & 72.0 & {\bf 74.8} & {\bf 60.2} & {\bf 68.8}$_{{\bf \uparrow 4.0}}$ \\

\bottomrule
\end{tabular}

\label{table:ablation}
\end{center}
\vspace{-0.1in}
\end{table}

\begin{figure*}[!t]
\centering
\includegraphics[width=1.0\linewidth]{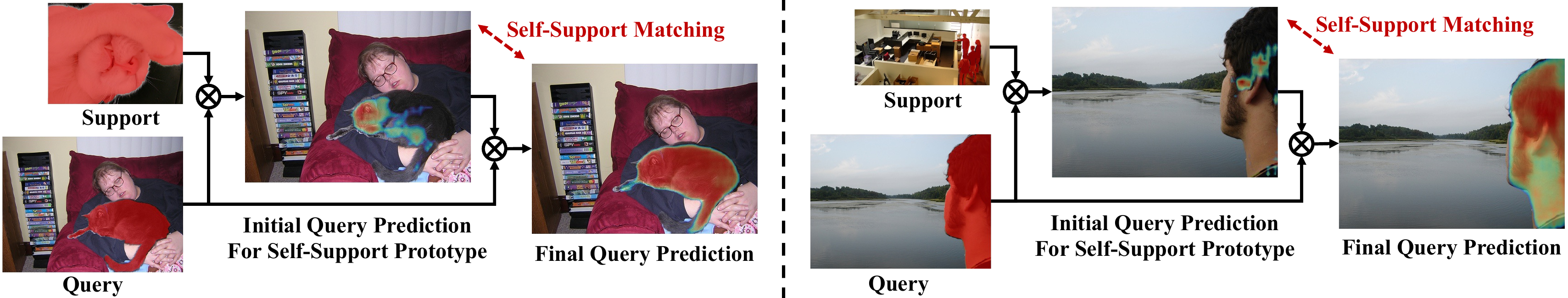}
\vspace{-0.3in}
\caption{Visualization for the working mechanism of our self-support matching. We omit the original support in self-support matching and the first row caption for clarity.}
\label{fig:vis}
\vspace{-0.1in}
\end{figure*}

\subsection{Self-Support Analysis}
We conduct extensive experiments and analysis to understand our method. 

\noindent{\bf Self-support working mechanism.~}
As shown in Figure~\ref{fig:vis}, we first generate the $\mathit{Initial}$ query predictions using the support prototype (as in Equation~\ref{eq:m1}), and leverage the confident predictions to extract query features to generate self-support prototype (as in Equation~\ref{eq:pq}). Then we use the self-support prototype to match with query features (as in Equation~\ref{eq:m2}) and produce the final output.
Note that because of the large inter-object/inter-background variation, the $\mathit{Init}$ predictions usually only capture some small representative regions, \eg, the cat/dog heads. Notwithstanding, our self-support method can handle well these hard cases by bridging the gap between query and support prototypes.

\begin{figure}[!t]
\centering
\includegraphics[width=1.0\linewidth]{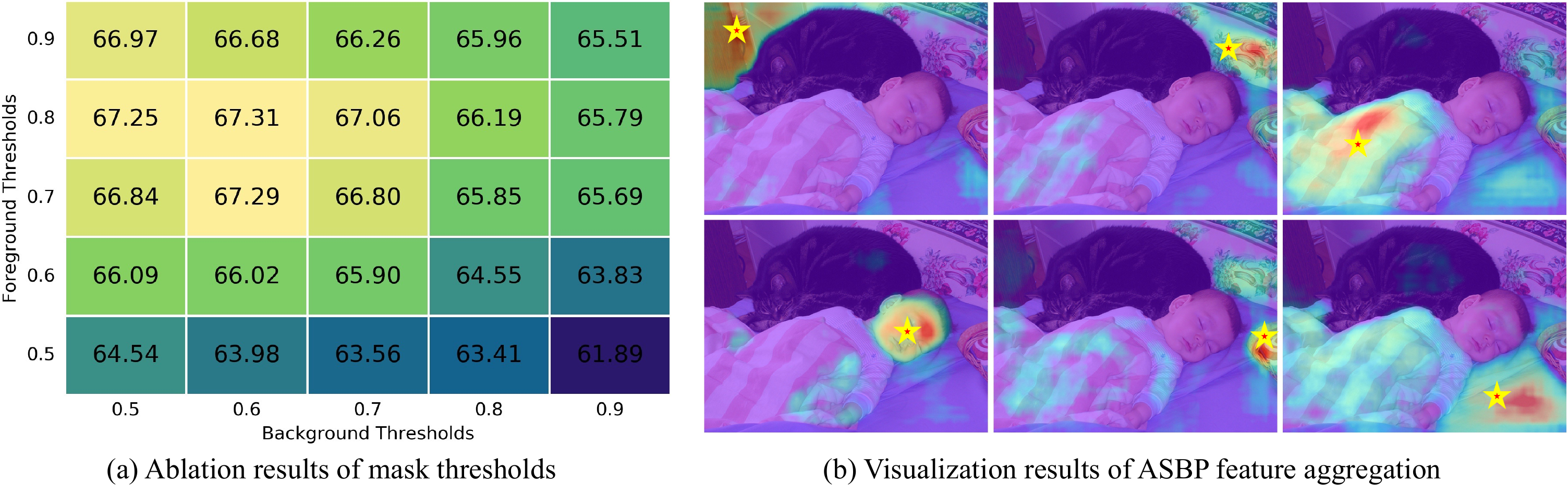}
\vspace{-0.3in}
\caption{(a) Results of mask threshold variations for self-support prototypes. (b) Visualization of the feature aggregation for adaptive self-support background prototypes (ASBP) at each star-marked position. They are aggregated from the activated background regions.}
\label{fig:threshold}
\vspace{-0.05in}
\end{figure}

\noindent{\bf Mask threshold.~} The threshold $\tau$ controls the query feature selection for self-support prototype generation (as in Equation~\ref{eq:pq}).
While we need to select high-confidence features for the foreground prototype,
the background prototype requires more query features with a relative low threshold.
This is because foreground pixels exhibit strong similarities 
with relatively low noise tolerance,
while background is cluttered and the aggregated diverse features should tolerate more noises.
Figure~\ref{fig:threshold} (a) summarizes the model performance on Pascal dataset with different thresholds, where a good balance between foreground and background thresholds are respectively
$\tau_{fg} \in [0.7, 0.9]$ and $\tau_{bg} \in [0.5, 0.7]$.

\noindent{\bf Prototype ablation.~} We investigate the effect of each of the prototypes by respectively removing them from the overall prototype. Table~\ref{table:prototype} summarizes
the results. Both our self-support foreground and background prototypes play a critical role to account for good matching performance. The support foreground prototype is also essential for the prototype quality thanks to its foreground semantic information aggregated from multiple support images. The support background prototype can be discarded with slight impact because of the large background variation between query and support.

\begin{table}[!t]
\begin{center}
\tabcolsep=8pt 

\caption{Ablation results of self-support module (SSM) by respectively removing foreground support prototype (FP), background support prototype (BP), self-support foreground prototype (SFP) and self-support background prototype (SBP).}
\vspace{-0.1in}
\begin{tabular}{ccccc}
\toprule
SSM & w/o FP & w/o BP & w/o SFP & w/o SBP \\
\midrule
67.3 & 66.0$_{\downarrow 1.3}$ & 67.2$_{\downarrow 0.1}$ & 66.5$_{\downarrow 0.8}$ & 65.6$_{\downarrow 1.7}$  \\
\bottomrule
\end{tabular}

\label{table:prototype}
\end{center}
\vspace{-0.3in}

\end{table}

\noindent{\bf Distinction from self-attention.~} Readers may compare 
our self-support method with self-attention mechanisms. Our self-support method shares some concepts but is different from self-attention.
Self-attention augments the image feature at each position by weighted aggregation of the features from all positions according to the affinity matrix.
In contrast, our self-support method leverages representative query features to generate prototypes according to the query-support matching results. 
In Table~\ref{table:self-attention} we experiment with multiple self-attention modules on the baseline. Unfortunately, all of them impose various degrees of harm on the matching performance, which are resulted by their self-attention augmentation which can destroy feature similarity between query and supports. 

\noindent{\bf Adaptive self-support background prototype.~}
This is designed to address the background clutter problem by adaptively aggregating background prototypes for each position.
As shown in Figure~\ref{fig:threshold} (b), the target cat is lying on a cluttered background consisting of the wardrobe, bed, quilt, baby, pillow and sheet. For each star-marked query position, the self-support background prototypes are aggregated from the corresponding semantic regions.
Note that this adaptive background prototype generation is specifically designed for self-support prototypes, which cannot be directly applied to support prototype generation because it will collapse to trivial solutions by greedily aggregating similar pixels without semantic consideration.


\begin{table}[!t]
\begin{center}
\tabcolsep=8pt 
\caption{Comparison with self-attention modules. ``$^\dagger$'' means the improved version by removing the transformation layer.}
\vspace{-0.1in}

\begin{tabular}{ccccc}
\toprule
Baseline & NL~\cite{wang2018non} & NL$^\dagger$~\cite{wang2018non} & GCNet~\cite{cao2019gcnet} & Our SSM \\
\midrule
64.8 & 62.1$_{\downarrow 2.7}$ & 64.3$_{\downarrow 0.5}$ & 63.9$_{\downarrow 0.9}$ & {\bf 67.3$_{{\bf \uparrow 2.5}}$} \\
\bottomrule
\end{tabular}

\label{table:self-attention}
\end{center}
\vspace{-0.2in}

\end{table}

\begin{table}[!t]
\begin{center}
\tabcolsep=8pt 
\caption{Comparison with other methods on performance improvement across different backbones and support shots.}
\vspace{-0.1in}

\begin{tabular}{cccccc}
\toprule
 &  PFENet~\cite{tian2020prior} & ReRPI~\cite{boudiaf2021few} & CWT~\cite{lu2021simpler} & MLC~\cite{yang2021mining} & Ours \\
\midrule
R50 $\rightarrow$ R101 & $-$0.5 & $-$1.2 & $+$1.0 & $+$2.7 & {\bf $+$3.7} \\
1shot $\rightarrow$ 5shot & $+$1.3 & $+$6.2 & $+$6.7 & $+$6.2 & {\bf $+$8.5} \\
\bottomrule
\end{tabular}

\label{table:benefit}
\end{center}
\vspace{-0.3in}

\end{table}

\subsection{Self-Support Advantages}
Our self-support method has many desirable properties.

\noindent{\bf Benefits from backbones and supports.~}
As shown before, our self-support method benefits more improvement from stronger backbones and more supports. 
Table~\ref{table:benefit} summarizes the improvement of different methods. When switching the backbone from ResNet-50 to ResNet-101, our method obtains 3.7\% performance improvement, while other methods obtain at most 2.7\% improvement or even performance degradation.
Our method also obtains the largest improvement of 8.5\% by increasing support images from 1-shot to 5-shot.
The behind reason is that our self-support method benefits from the Matthew effect~\cite{merton1968matthew} of accumulated advantages,
where better predictions induce better self-support prototypes and produce better predictions.

\noindent{\bf High-confident predictions.~}
Our self-support method not only improves  hard segmentation results with 0-1 labels, but also improves the soft confidence scores to produce high-confident predictions. As shown in Table~\ref{table:mse}, our self-support method significantly reduces the Mean Absolute Error (MAE) by 4.9\% compared to the baseline. We further evaluate the MAE in the truth positive (TP) regions for a fair comparison, where the MAE can still be largely reduced by 5.0\%.
These results demonstrate that our self-support method can significantly improve the output quality by producing high-confident predictions, a desirable property for many real-world applications.

\noindent{\bf Robust to weak support labels.~}
As shown in Table~\ref{table:weak}, when replacing the support mask with bounding box or scribble annotations for prototype generation, our self-support method still works very well with high robustness against support noises. This is because our method mainly relies on  self-support prototypes and thus is less affected by  
from noisy support prototypes.

\noindent{\bf Generalized to other methods.~}
Our self-support method is general and can be applied to other methods.
As shown in Table~\ref{table:others}, equipped with our self-support module, both the strong PANet~\cite{wang2019panet} and PPNet~\cite{liu2020part} report further boost in their performance by a large improvement.

\noindent{\bf High efficiency.~} Our self-support method is very efficient, which is a non-parametric method with few extra computation and $\sim$28 FPS running speed on a Tesla V100 GPU (with the ResNet-50 backbone in the 1-shot setting). 

\begin{table}[!t]
\begin{center}
\tabcolsep=8pt 
\caption{Results of prediction quality in MAE ($\downarrow$) metric. ``All/TP'' means evaluating models on all/truth positive regions of the image.}
\vspace{-0.1in}

\begin{tabular}{cccccc}
\toprule
 & Baseline & SSM & SSM+SSL & SSM+ASBP & Full \\
\midrule
All & 17.6 & 14.6$_{\downarrow 3.0}$ & 14.8$_{\downarrow 2.8}$ & 12.9$_{\downarrow 4.7}$ & {\bf 12.7$_{{\bf \downarrow 4.9}}$}  \\
TP & 13.2 & 9.6$_{\downarrow 3.6}$ & 10.1$_{\downarrow 3.1}$ & {\bf 7.8$_{{\bf \downarrow 5.4}}$} & 8.2$_{\downarrow 5.0}$  \\

\bottomrule
\end{tabular}

\label{table:mse}
\end{center}
\vspace{-0.2in}

\end{table}

\begin{table}[!t]
\begin{center}
\tabcolsep=8pt 
\caption{Results of using weak support annotations.}
\vspace{-0.1in}

\begin{tabular}{cccc}
\toprule
         & Mask & Scribble & Bounding Box \\
\midrule
Baseline & 64.8 & 63.3$_{\downarrow 1.5}$ & 61.7$_{\downarrow 3.1}$ \\
Ours     & 68.8 & 68.0$_{\downarrow 0.8}$ & 66.9$_{\downarrow 2.1}$ \\

\bottomrule
\end{tabular}

\label{table:weak}
\end{center}
\vspace{-0.2in}

\end{table}

\begin{table}[!t]
\begin{center}
\tabcolsep=8pt 
\caption{Results of applying our method to other models.}
\vspace{-0.1in}

\begin{tabular}{cc|cc}
\toprule
PANet~\cite{wang2019panet} & PANet + Ours & PPNet~\cite{liu2020part} & PPNet + Ours \\
\midrule
55.7 & 58.3$_{\uparrow 2.6}$ & 62.0 & 64.2$_{\uparrow 2.2}$ \\
\bottomrule
\end{tabular}

\label{table:others}
\end{center}

\vspace{-0.3in}
\end{table}

\section{Conclusion}

In this paper, we address the critical intra-class appearance discrepancy problem inherent in few-shot segmentation, 
by leveraging the query feature to generate self-support prototypes and perform self-support matching with query features.  
This strategy effectively narrows down the gap between support prototypes and query features. 
Further, we propose an adaptive self-support background prototype and a self-support loss to facilitate the self-support procedure.
Our self-support network has various desirable properties, and achieves SOTA on multiple benchmarks.
We have thoroughly investigated the self-support procedure with extensive experiments and analysis to substantiate its effectiveness and deepen our understanding on its working mechanism.

\section{More Implementation Details}

Our baseline model is adopted from MLC~\cite{yang2021mining} with a metric learning framework consisting of only an encoder.\blfootnote{This research was supported by Kuaishou Technology, the Research Grant Council of the HK SAR under grant No.~16201420, and NSFC fund (U2013210, 62006060).} \blfootnote{$^\dagger$ Corresponding author.}

Our improved model with self-support refinement is to repeat the self-support procedure based on the predicted mask $\mathcal{M}_2$ produced by our self-support network.
Specifically, the refined self-support foreground prototype $\mathcal{P}_{q,f}^r$ generation can be formulated as:
\begin{equation}
    \mathcal{P}_{q,f}^r = \mathit{MAP}(\widetilde{\mathcal{M}}_{2,q,f}, \mathcal{F}_q),
\end{equation}
where $\mathcal{F}_q$ is the query feature.
Similarly, we can generate the refined self-support background prototype $P_{q,b}^{\star,r}$ by 
\begin{equation}
    \mathcal{P}_{q,b}^{\star,r} = \mathit{ASBP}(\widetilde{\mathcal{M}}_{2,q,b}, \mathcal{F}_q),
\end{equation}
where $\mathit{ASBP}$ is the adaptive self-support background prototype generation module.
The $\widetilde{\mathcal{M}}_{2,q,f}=\mathbbm{1}(\mathcal{M}_{2,f} > \tau_{fg})$ and $\widetilde{\mathcal{M}}_{2,q,b}=\mathbbm{1}(\mathcal{M}_{2,b} > \tau_{bg})$ are defined according to the estimated query mask $\mathcal{M}_2 = \{\mathcal{M}_{2,f}, \mathcal{M}_{2,b}\}$ by Equation 8 in the main paper. We use the same $\{\tau_{fg}=0.7, \tau_{bg}=0.6\}$ settings as in the main paper.

Finally, we weight-combine the original support prototype $\mathcal{P}_s=\{\mathcal{P}_{s,f}, \mathcal{P}_{s,b}\}$, self-support prototype $\mathcal{P}_q=\{\mathcal{P}_{q,f}, \mathcal{P}_{q,b}^\star\}$ and refined self-support prototype $\mathcal{P}_q^r=\{\mathcal{P}_{q,f}^r, \mathcal{P}_{q,b}^{\star,r}\}$: 
\begin{equation}
    \mathcal{P}_s^{\star,r} = \alpha_1 \mathcal{P}_s + \alpha_2 \mathcal{P}_q + \alpha_3 \mathcal{P}_q^r,
\end{equation}
where $\alpha_1$, $\alpha_2$ and $\alpha_3$ are the tuning weights which are set as $\alpha_1 = 0.5$, $\alpha_2 = 0.2$ and $\alpha_3 = 0.3$ in our experiments.
Then we compute the cosine distance between the augmented support prototype with self-support refinement $\mathcal{P}_s^{\star,r}$ and query feature $\mathcal{F}_q$ to generate the matching prediction output $\mathcal{M}_3$:
\begin{equation}
\label{eq:m2}
\mathcal{M}_3 = \text{softmax}(\text{cosine}(\mathcal{P}_s^{\star,r}, \mathcal{F}_q)).
\end{equation}

The final output $M_{final}$ is the weighted combination of $\mathcal{M}_2$ and $\mathcal{M}_3$ for good performance:
\begin{equation}
\label{eq:m2}
\mathcal{M}_{final} = \beta_1 M_2 + \beta_2 M_3,
\end{equation}
where $\beta_1$ and $\beta_2$ are the tuning weights and we set $\beta_1 = 0.3$ and $\beta_2 = 0.7$ in our experiments.

\section{More Quantitative Results}

In the main paper, we repeat the evaluation procedure of all our experiments by 5 times with different random seeds to obtain stable results.

To further validate the effectiveness of our self-support method, we evaluate our model on FSS-1000~\cite{li2020fss}, which is a recently proposed large-scale few-shot segmentation dataset containing 1000 classes. The dataset is split into train/val/test sets with 520, 240, 240 classes respectively. We follow the common practice~\cite{li2020fss} to evaluate performance on FSS-1000 using the intersection-over-union (IoU) of positive labels in a binary segmentation map. The evaluation procedure is conducted on 2400 randomly sampled support-query pairs. As shown in Table~\ref{table:fss-1000}, our self-support matching model outperforms other methods. And the self-support refinement step can further promote our performance to 87.3/88.6 mIoU in 1/5-shot settings. 


As shown in Table~\ref{table:improvement}, we present the performance improvement on Pascal VOC dataset~\cite{everingham2010pascal} of our self-support method on the baseline models. Our method can consistently improve the performance by a large margin with different backbone models and support shots. We can also observe more performance gains on the stronger backbone model and more support shots, which is consistent with the advantage conclusion in the main paper.


\begin{table}[!t]
\begin{center}
\tabcolsep=8pt 
\caption{Quantitative comparison results on FSS-1000 dataset with the mIoU metric of positive labels in a binary segmentation map.}
\begin{tabular}{cccc}
\toprule
Method & Publication & 1-shot & 5-shot \\
\midrule
OSLSM~\cite{shaban2017one} & BMVC'17 & 70.3 & 73.0 \\
GNet~\cite{rakelly2018few} & Arxiv'18 & 71.9 & 74.3 \\
FSS~\cite{li2020fss} & CVPR'20 & 73.5 & 80.1 \\
DoG-LSTM~\cite{azad2021texture} & WACV'21 & 80.8 & 83.4 \\
DAN~\cite{wang2020few} & ECCV'20 & 85.2 & 88.1 \\
SSP (Ours) & - & 86.9 & 88.2 \\
SSP$_{\mathit{refine}}$ & - & 87.3 & 88.6 \\

\bottomrule
\end{tabular}
\label{table:fss-1000}
\end{center}
\end{table}

\begin{table}[!t]
\begin{center}
\tabcolsep=8pt 
\caption{Performance improvement on Pascal VOC dataset of our self-support method on baseline models with different backbones and support shots.}

\begin{tabular}{ccccc}
\toprule
Backbone & Shot & Baseline & Ours & $\Delta$ \\
\midrule
\multirow{2}{*}{ResNet-50} & 1-shot & 57.8 & 60.9 & +3.1 \\
 & 5-shot & 64.8 & 68.8 & +4.0 \\
\midrule
\multirow{2}{*}{ResNet-101} & 1-shot & 60.1 & 64.0 & +3.9 \\
 & 5-shot & 67.8 & 72.5 & +4.7 \\


\bottomrule
\end{tabular}
\label{table:improvement}
\end{center}
\end{table}

\section{More Qualitative Results}

\begin{figure*}[!t]
\centering
\includegraphics[width=1.0\linewidth]{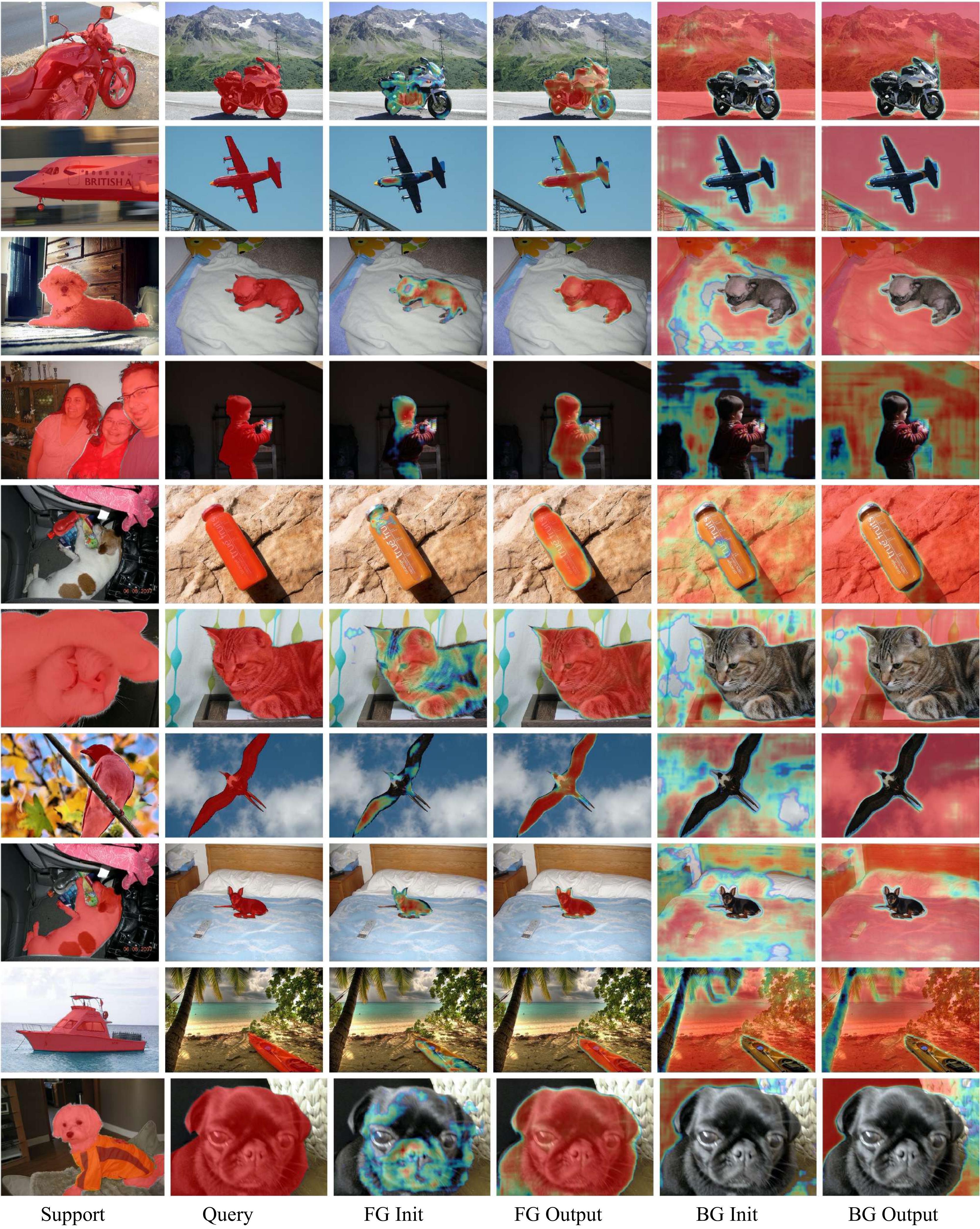}
\caption{The 1-shot visualization results of our model containing the \textit{Init} and final outputs.}
\label{fig:supp_1}
\end{figure*}

\begin{figure*}[!t]
\centering
\includegraphics[width=1.0\linewidth]{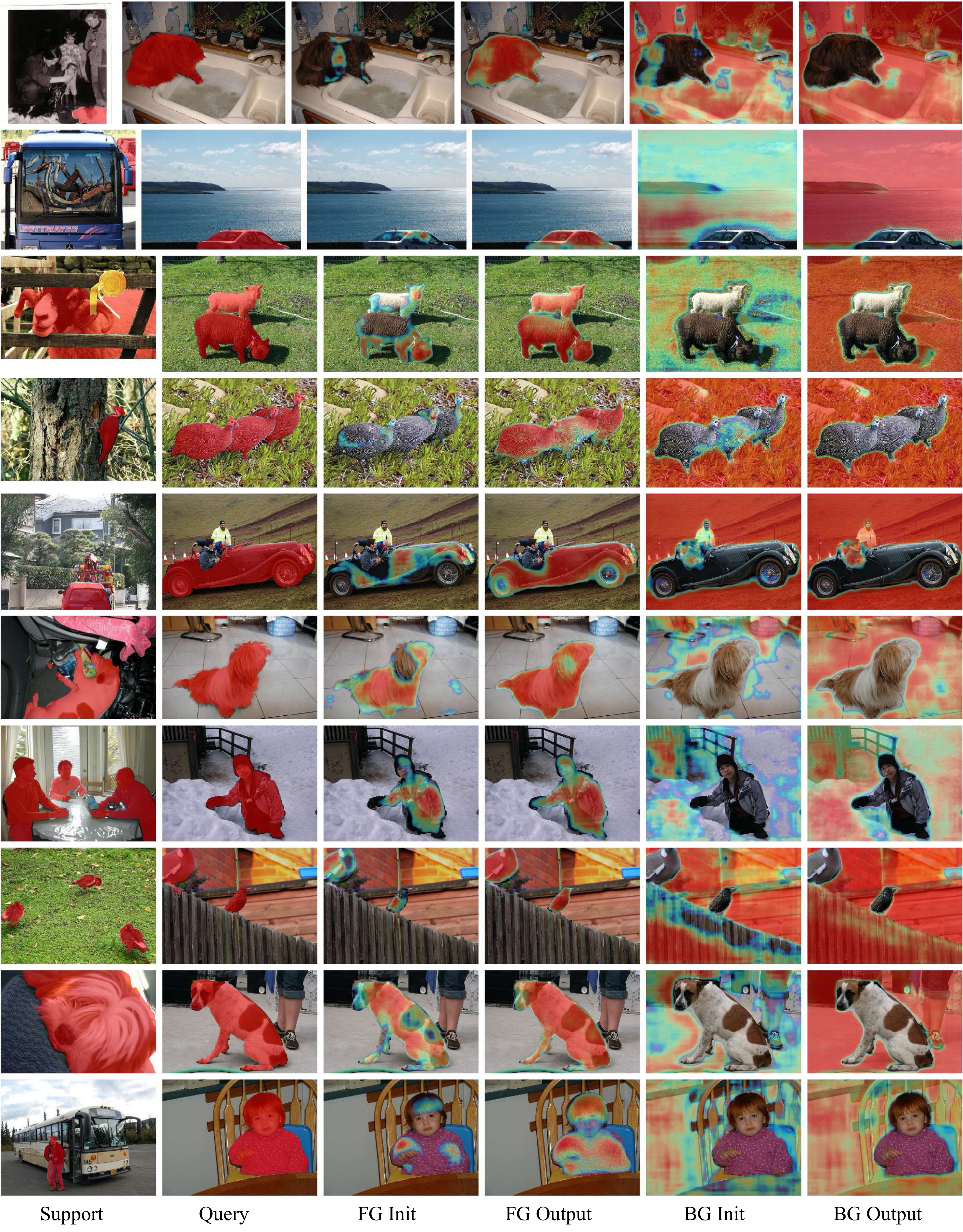}
\caption{The 1-shot visualization results of our model containing the \textit{Init} and final outputs.}
\label{fig:supp_2}
\end{figure*}

\begin{figure*}[!t]
\centering
\includegraphics[width=1.0\linewidth]{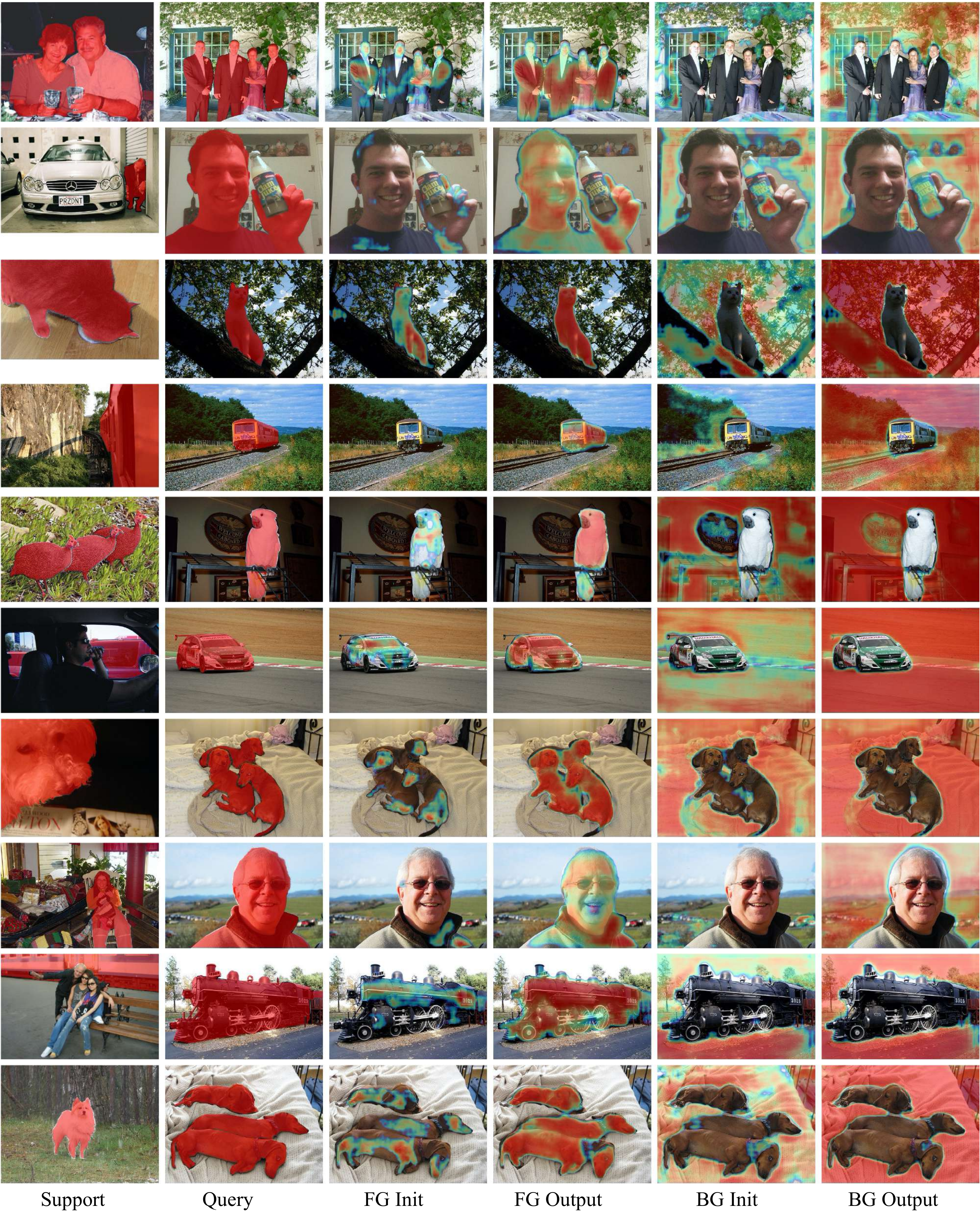}
\caption{The 1-shot visualization results of our model containing the \textit{Init} and final outputs.}
\label{fig:supp_3}
\end{figure*}

\begin{figure*}[!t]
\centering
\includegraphics[width=1.0\linewidth]{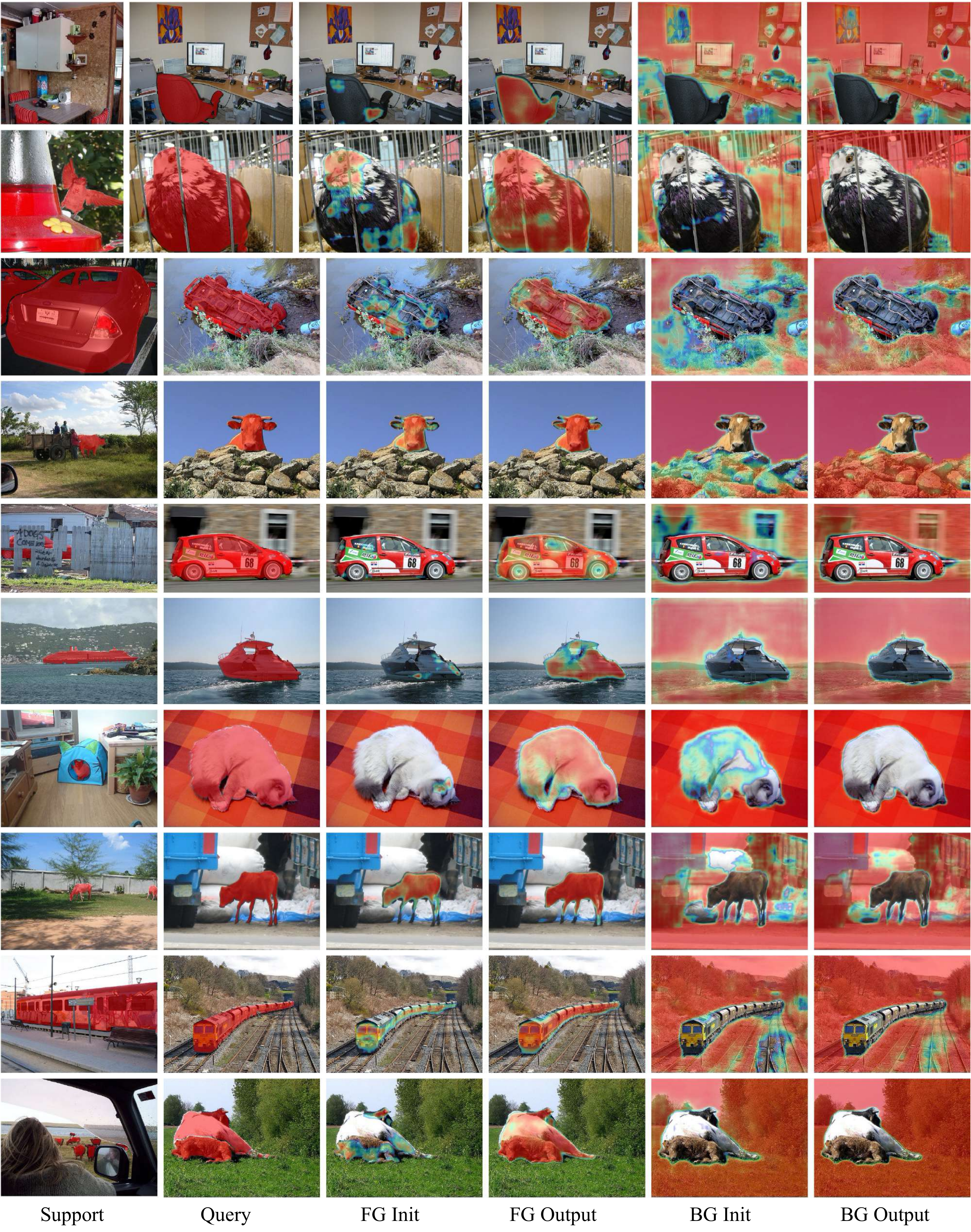}
\caption{The 1-shot visualization results of our model containing the \textit{Init} and final outputs.}
\label{fig:supp_4}
\end{figure*}

We present more qualitative results in 1-shot setting with ResNet-50 backbone for better visualization. As shown in Figure~\ref{fig:supp_1}, Figure~\ref{fig:supp_2}, Figure~\ref{fig:supp_3}, and Figure~\ref{fig:supp_4}, objects in support and query images have large appearance discrepancy even belonging to the same class.
Thus the initial predictions generated by the traditional matching network can cover only a small region of the target object. On the other hand, equipped with our self-support method, the model can produce satisfactory results with substantial qualitative improvement. Note that the initial masks are obtained by setting foreground and background thresholds on the original mask prediction $\mathcal{M}_1$.

\clearpage

\clearpage

\clearpage

\clearpage

%
%
\bibliographystyle{splncs04}
\bibliography{main}

\begin{thebibliography}{10}
\providecommand{\url}[1]{\texttt{#1}}
\providecommand{\urlprefix}{URL }
\providecommand{\doi}[1]{https://doi.org/#1}

\bibitem{allen2019infinite}
Allen, K., Shelhamer, E., Shin, H., Tenenbaum, J.: Infinite mixture prototypes
  for few-shot learning. In: ICML (2019)

\bibitem{antoniou2018train}
Antoniou, A., Edwards, H., Storkey, A.: How to train your maml. In: ICLR (2019)

\bibitem{azad2021texture}
Azad, R., Fayjie, A.R., Kauffmann, C., Ben~Ayed, I., Pedersoli, M., Dolz, J.:
  On the texture bias for few-shot cnn segmentation. In: WACV (2021)

\bibitem{badrinarayanan2017segnet}
Badrinarayanan, V., Kendall, A., Cipolla, R.: Segnet: A deep convolutional
  encoder-decoder architecture for image segmentation. IEEE TPAMI  (2017)

\bibitem{benenson2019large}
Benenson, R., Popov, S., Ferrari, V.: Large-scale interactive object
  segmentation with human annotators. In: CVPR (2019)

\bibitem{bertinetto2018meta}
Bertinetto, L., Henriques, J.F., Torr, P.H., Vedaldi, A.: Meta-learning with
  differentiable closed-form solvers. In: ICLR (2019)

\bibitem{boudiaf2021few}
Boudiaf, M., Kervadec, H., Masud, Z.I., Piantanida, P., Ben~Ayed, I., Dolz, J.:
  Few-shot segmentation without meta-learning: A good transductive inference is
  all you need? In: CVPR (2021)

\bibitem{cao2019gcnet}
Cao, Y., Xu, J., Lin, S., Wei, F., Hu, H.: Gcnet: Non-local networks meet
  squeeze-excitation networks and beyond. In: CVPRW (2019)

\bibitem{chen2017deeplab}
Chen, L.C., Papandreou, G., Kokkinos, I., Murphy, K., Yuille, A.L.: Deeplab:
  Semantic image segmentation with deep convolutional nets, atrous convolution,
  and fully connected crfs. IEEE TPAMI  (2017)

\bibitem{chen2016attention}
Chen, L.C., Yang, Y., Wang, J., Xu, W., Yuille, A.L.: Attention to scale:
  Scale-aware semantic image segmentation. In: CVPR (2016)

\bibitem{chen2018encoder}
Chen, L.C., Zhu, Y., Papandreou, G., Schroff, F., Adam, H.: Encoder-decoder
  with atrous separable convolution for semantic image segmentation. In: ECCV
  (2018)

\bibitem{chen2019closer}
Chen, W.Y., Liu, Y.C., Kira, Z., Wang, Y.C.F., Huang, J.B.: A closer look at
  few-shot classification. In: ICLR (2019)

\bibitem{cheng2019spgnet}
Cheng, B., Chen, L.C., Wei, Y., Zhu, Y., Huang, Z., Xiong, J., Huang, T.S.,
  Hwu, W.M., Shi, H.: Spgnet: Semantic prediction guidance for scene parsing.
  In: ICCV (2019)

\bibitem{dai2017deformable}
Dai, J., Qi, H., Xiong, Y., Li, Y., Zhang, G., Hu, H., Wei, Y.: Deformable
  convolutional networks. In: ICCV (2017)

\bibitem{deng2009imagenet}
Deng, J., Dong, W., Socher, R., Li, L.J., Li, K., Fei-Fei, L.: Imagenet: A
  large-scale hierarchical image database. In: CVPR (2009)

\bibitem{dhillon2019baseline}
Dhillon, G.S., Chaudhari, P., Ravichandran, A., Soatto, S.: A baseline for
  few-shot image classification. In: ICLR (2019)

\bibitem{doersch2020crosstransformers}
Doersch, C., Gupta, A., Zisserman, A.: Crosstransformers: spatially-aware
  few-shot transfer. In: NeurIPS (2020)

\bibitem{dong2018few}
Dong, N., Xing, E.P.: Few-shot semantic segmentation with prototype learning.
  In: BMVC (2018)

\bibitem{everingham2010pascal}
Everingham, M., Van~Gool, L., Williams, C.K., Winn, J., Zisserman, A.: The
  pascal visual object classes (voc) challenge. IJCV  (2010)

\bibitem{fan2021group}
Fan, Q., Fan, D.P., Fu, H., Tang, C.K., Shao, L., Tai, Y.W.: Group
  collaborative learning for co-salient object detection. In: CVPR (2021)

\bibitem{fan2020cpmask}
Fan, Q., Ke, L., Pei, W., Tang, C.K., Tai, Y.W.: Commonality-parsing network
  across shape and appearance for partially supervised instance segmentation.
  In: ECCV (2020)

\bibitem{fan2021few}
Fan, Q., Tang, C.K., Tai, Y.W.: Few-shot video object detection. arXiv preprint
  arXiv:2104.14805  (2021)

\bibitem{fan2020few}
Fan, Q., Zhuo, W., Tang, C.K., Tai, Y.W.: Few-shot object detection with
  attention-rpn and multi-relation detector. In: CVPR (2020)

\bibitem{Finn2017ModelAgnosticMF}
Finn, C., Abbeel, P., Levine, S.: Model-agnostic meta-learning for fast
  adaptation of deep networks. In: ICML (2017)

\bibitem{fu2019dual}
Fu, J., Liu, J., Tian, H., Li, Y., Bao, Y., Fang, Z., Lu, H.: Dual attention
  network for scene segmentation. In: CVPR (2019)

\bibitem{fu2019adaptive}
Fu, J., Liu, J., Wang, Y., Li, Y., Bao, Y., Tang, J., Lu, H.: Adaptive context
  network for scene parsing. In: ICCV (2019)

\bibitem{gairola2020simpropnet}
Gairola, S., Hemani, M., Chopra, A., Krishnamurthy, B.: Simpropnet: Improved
  similarity propagation for few-shot image segmentation. In: IJCAI (2020)

\bibitem{gidaris2018dynamic}
Gidaris, S., Komodakis, N.: Dynamic few-shot visual learning without
  forgetting. In: CVPR (2018)

\bibitem{Goodfellow-et-al-2016}
Goodfellow, I., Bengio, Y., Courville, A.: Deep Learning. MIT Press (2016)

\bibitem{gordon2018metalearning}
Gordon, J., Bronskill, J., Bauer, M., Nowozin, S., Turner, R.: Meta-learning
  probabilistic inference for prediction. In: ICLR (2019)

\bibitem{grant2018recasting}
Grant, E., Finn, C., Levine, S., Darrell, T., Griffiths, T.: Recasting
  gradient-based meta-learning as hierarchical bayes. In: ICLR (2018)

\bibitem{he2021progressive}
He, H., Zhang, J., Thuraisingham, B., Tao, D.: Progressive one-shot human
  parsing. In: AAAI (2021)

\bibitem{he2019adaptive}
He, J., Deng, Z., Zhou, L., Wang, Y., Qiao, Y.: Adaptive pyramid context
  network for semantic segmentation. In: CVPR (2019)

\bibitem{he2016deep}
He, K., Zhang, X., Ren, S., Sun, J.: Deep residual learning for image
  recognition. In: CVPR (2016)

\bibitem{hou2019cross}
Hou, R., Chang, H., Ma, B., Shan, S., Chen, X.: Cross attention network for
  few-shot classification. In: NeurIPS (2019)

\bibitem{huang2019ccnet}
Huang, Z., Wang, X., Huang, L., Huang, C., Wei, Y., Liu, W.: Ccnet: Criss-cross
  attention for semantic segmentation. In: ICCV (2019)

\bibitem{kang2018few}
Kang, B., Liu, Z., Wang, X., Yu, F., Feng, J., Darrell, T.: Few-shot object
  detection via feature reweighting. In: ICCV (2019)

\bibitem{kim2021uncertainty}
Kim, S., Chikontwe, P., Park, S.H.: Uncertainty-aware semi-supervised few shot
  segmentation. In: IJCAI (2021)

\bibitem{kirillov2019panoptic}
Kirillov, A., Girshick, R., He, K., Doll{\'a}r, P.: Panoptic feature pyramid
  networks. In: CVPR (2019)

\bibitem{koch2015siamese}
Koch, G., Zemel, R., Salakhutdinov, R.: Siamese neural networks for one-shot
  image recognition. In: ICMLW (2015)

\bibitem{gestalt}
Koffka, K.: Principles of Gestalt psychology. Routledge (1935)

\bibitem{krizhevsky2012imagenet}
Krizhevsky, A., Sutskever, I., Hinton, G.E.: Imagenet classification with deep
  convolutional neural networks. In: NeurIPS (2012)

\bibitem{lee2019meta}
Lee, K., Maji, S., Ravichandran, A., Soatto, S.: Meta-learning with
  differentiable convex optimization. In: CVPR (2019)

\bibitem{lee2018gradient}
Lee, Y., Choi, S.: Gradient-based meta-learning with learned layerwise metric
  and subspace. In: ICML (2018)

\bibitem{li2021adaptive}
Li, G., Jampani, V., Sevilla-Lara, L., Sun, D., Kim, J., Kim, J.: Adaptive
  prototype learning and allocation for few-shot segmentation. In: CVPR (2021)

\bibitem{li2019finding}
Li, H., Eigen, D., Dodge, S., Zeiler, M., Wang, X.: Finding task-relevant
  features for few-shot learning by category traversal. In: CVPR (2019)

\bibitem{li2019revisiting}
Li, W., Wang, L., Xu, J., Huo, J., Gao, Y., Luo, J.: Revisiting local
  descriptor based image-to-class measure for few-shot learning. In: CVPR
  (2019)

\bibitem{li2020fss}
Li, X., Wei, T., Chen, Y.P., Tai, Y.W., Tang, C.K.: Fss-1000: A 1000-class
  dataset for few-shot segmentation. In: CVPR (2020)

\bibitem{lin2017refinenet}
Lin, G., Milan, A., Shen, C., Reid, I.: Refinenet: Multi-path refinement
  networks for high-resolution semantic segmentation. In: CVPR (2017)

\bibitem{lin2016efficient}
Lin, G., Shen, C., Van Den~Hengel, A., Reid, I.: Efficient piecewise training
  of deep structured models for semantic segmentation. In: CVPR (2016)

\bibitem{lin2014microsoft}
Lin, T.Y., Maire, M., Belongie, S., Hays, J., Perona, P., Ramanan, D.,
  Doll{\'a}r, P., Zitnick, C.L.: Microsoft coco: Common objects in context. In:
  ECCV (2014)

\bibitem{liu2021anti}
Liu, B., Ding, Y., Jiao, J., Ji, X., Ye, Q.: Anti-aliasing semantic
  reconstruction for few-shot semantic segmentation. In: CVPR (2021)

\bibitem{liu2021learning}
Liu, C., Fu, Y., Xu, C., Yang, S., Li, J., Wang, C., Zhang, L.: Learning a
  few-shot embedding model with contrastive learning. In: AAAI (2021)

\bibitem{liu2020dynamic}
Liu, L., Cao, J., Liu, M., Guo, Y., Chen, Q., Tan, M.: Dynamic extension nets
  for few-shot semantic segmentation. In: ACM MM (2020)

\bibitem{liu2020crnet}
Liu, W., Zhang, C., Lin, G., Liu, F.: Crnet: Cross-reference networks for
  few-shot segmentation. In: CVPR (2020)

\bibitem{liu2020part}
Liu, Y., Zhang, X., Zhang, S., He, X.: Part-aware prototype network for
  few-shot semantic segmentation. In: ECCV (2020)

\bibitem{liu2020ppnet}
Liu, Y., Zhang, X., Zhang, S., He, X.: Part-aware prototype network for
  few-shot semantic segmentation. In: ECCV (2020)

\bibitem{long2015fully}
Long, J., Shelhamer, E., Darrell, T.: Fully convolutional networks for semantic
  segmentation. In: CVPR (2015)

\bibitem{lu2021simpler}
Lu, Z., He, S., Zhu, X., Zhang, L., Song, Y.Z., Xiang, T.: Simpler is better:
  Few-shot semantic segmentation with classifier weight transformer. In: ICCV
  (2021)

\bibitem{merton1968matthew}
Merton, R.K.: The matthew effect in science: The reward and communication
  systems of science are considered. Science  (1968)

\bibitem{min2021hypercorrelation}
Min, J., Kang, D., Cho, M.: Hypercorrelation squeeze for few-shot segmentation.
  In: ICCV (2021)

\bibitem{nguyen2019feature}
Nguyen, K., Todorovic, S.: Feature weighting and boosting for few-shot
  segmentation. In: ICCV (2019)

\bibitem{noh2015learning}
Noh, H., Hong, S., Han, B.: Learning deconvolution network for semantic
  segmentation. In: ICCV (2015)

\bibitem{ouyang2020self}
Ouyang, C., Biffi, C., Chen, C., Kart, T., Qiu, H., Rueckert, D.:
  Self-supervision with superpixels: Training few-shot medical image
  segmentation without annotation. In: ECCV (2020)

\bibitem{qi2018low}
Qi, H., Brown, M., Lowe, D.G.: Low-shot learning with imprinted weights. In:
  CVPR (2018)

\bibitem{rakelly2018few}
Rakelly, K., Shelhamer, E., Darrell, T., Efros, A.A., Levine, S.: Few-shot
  segmentation propagation with guided networks. arXiv preprint
  arXiv:1806.07373  (2018)

\bibitem{ronneberger2015u}
Ronneberger, O., Fischer, P., Brox, T.: U-net: Convolutional networks for
  biomedical image segmentation. In: MICCAI (2015)

\bibitem{rusu2018meta}
Rusu, A.A., Rao, D., Sygnowski, J., Vinyals, O., Pascanu, R., Osindero, S.,
  Hadsell, R.: Meta-learning with latent embedding optimization. In: ICLR
  (2019)

\bibitem{shaban2017one}
Shaban, A., Bansal, S., Liu, Z., Essa, I., Boots, B.: One-shot learning for
  semantic segmentation. In: BMVC (2017)

\bibitem{siam2020weakly}
Siam, M., Doraiswamy, N., Oreshkin, B.N., Yao, H., Jagersand, M.: Weakly
  supervised few-shot object segmentation using co-attention with visual and
  semantic embeddings. In: IJCAI (2020)

\bibitem{siam2019amp}
Siam, M., Oreshkin, B.N., Jagersand, M.: Amp: Adaptive masked proxies for
  few-shot segmentation. In: ICCV (2019)

\bibitem{tian2020differentiable}
Tian, P., Wu, Z., Qi, L., Wang, L., Shi, Y., Gao, Y.: Differentiable
  meta-learning model for few-shot semantic segmentation. In: AAAI (2020)

\bibitem{tian2020prior}
Tian, Z., Zhao, H., Shu, M., Yang, Z., Li, R., Jia, J.: Prior guided feature
  enrichment network for few-shot segmentation. IEEE TPAMI  (2020)

\bibitem{wang2020few}
Wang, H., Zhang, X., Hu, Y., Yang, Y., Cao, X., Zhen, X.: Few-shot semantic
  segmentation with democratic attention networks. In: ECCV (2020)

\bibitem{wang2019panet}
Wang, K., Liew, J.H., Zou, Y., Zhou, D., Feng, J.: Panet: Few-shot image
  semantic segmentation with prototype alignment. In: ICCV (2019)

\bibitem{wang2018non}
Wang, X., Girshick, R., Gupta, A., He, K.: Non-local neural networks. In: CVPR
  (2018)

\bibitem{wu2021learning}
Wu, Z., Shi, X., Lin, G., Cai, J.: Learning meta-class memory for few-shot
  semantic segmentation. In: ICCV (2021)

\bibitem{xie2021scale}
Xie, G.S., Liu, J., Xiong, H., Shao, L.: Scale-aware graph neural network for
  few-shot semantic segmentation. In: CVPR (2021)

\bibitem{xie2021few}
Xie, G.S., Xiong, H., Liu, J., Yao, Y., Shao, L.: Few-shot semantic
  segmentation with cyclic memory network. In: ICCV (2021)

\bibitem{yan2019metarcnn}
Yan, X., Chen, Z., Xu, A., Wang, X., Liang, X., Lin, L.: Meta r-cnn : Towards
  general solver for instance-level low-shot learning. In: ICCV (2019)

\bibitem{yang2020prototype}
Yang, B., Liu, C., Li, B., Jiao, J., Ye, Q.: Prototype mixture models for
  few-shot semantic segmentation. In: ECCV (2020)

\bibitem{yang2021mining}
Yang, L., Zhuo, W., Qi, L., Shi, Y., Gao, Y.: Mining latent classes for
  few-shot segmentation. In: ICCV (2021)

\bibitem{yang2020brinet}
Yang, X., Wang, B., Chen, K., Zhou, X., Yi, S., Ouyang, W., Zhou, L.: Brinet:
  Towards bridging the intra-class and inter-class gaps in one-shot
  segmentation. In: BMVC (2020)

\bibitem{yu2017dilated}
Yu, F., Koltun, V., Funkhouser, T.: Dilated residual networks. In: CVPR (2017)

\bibitem{yuan2020object}
Yuan, Y., Chen, X., Wang, J.: Object-contextual representations for semantic
  segmentation. In: ECCV (2020)

\bibitem{zhang2021self}
Zhang, B., Xiao, J., Qin, T.: Self-guided and cross-guided learning for
  few-shot segmentation. In: CVPR (2021)

\bibitem{zhang2019pyramid}
Zhang, C., Lin, G., Liu, F., Guo, J., Wu, Q., Yao, R.: Pyramid graph networks
  with connection attentions for region-based one-shot semantic segmentation.
  In: ICCV (2019)

\bibitem{zhang2019canet}
Zhang, C., Lin, G., Liu, F., Yao, R., Shen, C.: Canet: Class-agnostic
  segmentation networks with iterative refinement and attentive few-shot
  learning. In: CVPR (2019)

\bibitem{zhang2019acfnet}
Zhang, F., Chen, Y., Li, Z., Hong, Z., Liu, J., Ma, F., Han, J., Ding, E.:
  Acfnet: Attentional class feature network for semantic segmentation. In: ICCV
  (2019)

\bibitem{zhang2020few}
Zhang, H., Zhang, L., Qi, X., Li, H., Torr, P.H., Koniusz, P.: Few-shot action
  recognition with permutation-invariant attention. In: ECCV (2020)

\bibitem{zhao2017pyramid}
Zhao, H., Shi, J., Qi, X., Wang, X., Jia, J.: Pyramid scene parsing network.
  In: CVPR (2017)

\bibitem{zhao2018psanet}
Zhao, H., Zhang, Y., Liu, S., Shi, J., Loy, C.C., Lin, D., Jia, J.: Psanet:
  Point-wise spatial attention network for scene parsing. In: ECCV (2018)

\bibitem{zhou2017scene}
Zhou, B., Zhao, H., Puig, X., Fidler, S., Barriuso, A., Torralba, A.: Scene
  parsing through ade20k dataset. In: CVPR (2017)

\bibitem{zhu2020self}
Zhu, K., Zhai, W., Zha, Z.J., Cao, Y.: Self-supervised tuning for few-shot
  segmentation. In: IJCAI (2020)

\bibitem{zhu2019asymmetric}
Zhu, Z., Xu, M., Bai, S., Huang, T., Bai, X.: Asymmetric non-local neural
  networks for semantic segmentation. In: ICCV (2019)

\bibitem{zhuge2021deep}
Zhuge, Y., Shen, C.: Deep reasoning network for few-shot semantic segmentation.
  In: ACM MM (2021)

\end{thebibliography}
\end{document}